\documentclass[10pt,letterpaper]{article}
\usepackage{cogsci}

\cogscifinalcopy % Uncomment this line for the final submission 

\usepackage{pslatex}
\usepackage{amsmath,amsfonts,bm}

\def\eqref#1{equation~\ref{#1}}
\def\1{\bm{1}}

\DeclareMathAlphabet{\mathsfit}{\encodingdefault}{\sfdefault}{m}{sl}
\SetMathAlphabet{\mathsfit}{bold}{\encodingdefault}{\sfdefault}{bx}{n}

\usepackage{hyperref}
\usepackage{url}
\usepackage{apacite}

\usepackage{booktabs}       % professional-quality tables
\usepackage{amsfonts}       % blackboard math symbols
\usepackage{nicefrac}       % compact symbols for 1/2, etc.
\usepackage{microtype}      % microtypography

\usepackage{algorithm}
\usepackage[noend]{algpseudocode}
\usepackage{amsmath}
\usepackage{graphicx}
\usepackage{subcaption}

\usepackage{framed}
\usepackage{comment}
\usepackage{array}
\newcolumntype{L}[1]{>{\raggedright\let\newline\\\arraybackslash\hspace{0pt}}m{#1}}

\usepackage{multirow}
\usepackage{makecell}
\usepackage{diagbox}

\usepackage[]{color}

\newcommand{\cob}{\color{black}}

\usepackage{comment}
\newcommand{\bco}{\begin{comment}}
\newcommand{\eco}{\end{comment}}

\title{Mental representations of objects reflect the ways in which we interact with them}

\newif\ifdefaultauthor
\defaultauthorfalse

\ifdefaultauthor

\author{{\large \bf Morton Ann Gernsbacher (MAG@Macc.Wisc.Edu)} \\
  Department of Psychology, 1202 W. Johnson Street \\
  Madison, WI 53706 USA
  \AND {\large \bf Sharon J.~Derry (SDJ@Macc.Wisc.Edu)} \\
  Department of Educational 
}

\else

\author{Ka Chun Lam$^1$, Francisco Pereira$^1$, Maryam Vaziri-Pashkam$^2$, Kristin Woodard$^2$, Emalie McMahon$^3$  \\
$^1$Machine Learning Team, $^2$Laboratory for Brain and Cognition\\
National Institute of Mental Health, Bethesda, MD 20892, USA \\
\texttt{\{kachun.lam, francisco.pereira, maryam.vaziri-pashkam, kristin.woodard\}@nih.gov} 
\AND
$^3$Department of Cognitive Science \\
Johns Hopkins University, Baltimore, MD 21218, USA \\
\texttt{emaliemcmahon@jhu.edu}
}
\fi

\begin{document}

\maketitle

\begin{abstract}

In order to interact with objects in our environment, humans rely on an understanding of the actions that can be performed on them, as well as their properties. When considering concrete motor actions, this knowledge has been called the object affordance. Can this notion be generalized to any type of interaction that one can have with an object? In this paper we introduce a method to represent objects in a space where each dimension corresponds to a broad mode of interaction, based on verb selectional preferences in text corpora. This object embedding makes it possible to predict human judgments of verb applicability to objects better than a variety of alternative approaches. Furthermore, we show that the dimensions in this space can be used to predict categorical and functional dimensions in a state-of-the-art mental representation of objects, derived solely from human judgements of object similarity. These results suggest that interaction knowledge accounts for a large part of mental representations of objects.

\end{abstract}

\section{Introduction}
 
In order to interact with objects in our environment, we rely on an understanding of the actions that can be performed on them, and their dependence (or effect) on properties of the object. \citeA{gibson1979ecological} coined the term ``affordance'' to describe what the environment ``provides or furnishes the animal''. \citeA{norman2013design} developed the term to focus on the properties of objects that determine action possibilities. The notion of ``affordance'' emerges from the relationship between the properties of objects and human actions.
If we consider ``object'' as meaning anything concrete that one might interact with in the environment, there will be thousands of possibilities, both animate and inanimate (see WordNet~\cite{miller1998wordnet}). The same is true if we consider ``action'' as meaning any verb that might be applied to the noun naming an object (see VerbNet~\cite{schuler2005verbnet}). Intuitively, only a relatively small fraction of all possible combinations of object and action will be plausible. Of those, many will also be trivial, e.g. ``see'' or ``have'' may apply to almost every object. Finally,  different actions might reflect a similar mode of interaction, depending on the type of object they are applied to (e.g. "chop" and "slice" are distinct actions, but they are both used in food preparation).

Mental representations of objects encompass many aspects beyond function. Several studies \cite{mcrae2005semantic,devereux2014centre,hovhannisyan2020visual} have normed thousands of binary properties for hundreds of objects.
%asked human subjects to list binary properties for hundreds of objects, yielding thousands of answers. 
Properties could be taxonomic (category), functional (purpose), encyclopedic (attributes), or visual-perceptual (appearance), among others. While some properties were affordances in themselves, most reflected many affordances at once (e.g. ``is a vegetable'' means that it could be planted, cooked, sliced, etc).

Recently, \citeA{ZhengPBH19} and \citeA{hebart2020revealing} introduced SPoSE, a model of the mental representations of 1,854 objects in a 49-dimensional space. The model was derived from a dataset of 1.5M Amazon Mechanical Turk (AMT) judgments of object similarity, where subjects were asked which of a random triplet of objects was the odd one out. The model embedded each object as a vector in a space where each dimension was constrained to be sparse and positive. Triplet judgments were predicted as a function of the similarity between embedding vectors of the three objects considered. The authors showed that these dimensions were predictable as a {\em combination} of elementary properties in the \cite{devereux2014centre} norm that often co-occur across many objects. \citeA{hebart2020revealing} further showed that 1) human subjects could coherently label what the dimensions were ``about'', ranging from categorical (e.g. is animate, food, drink, building) to functional (e.g. container, tool) or structural (e.g. made of metal or wood, has inner structure). Subjects could also predict what dimension values new objects would have, based on knowing the dimension value for a few other objects.

Our first goal is to produce an analogous "affordance embedding" for objects, where each dimension of the space groups together actions often applied to objects scoring high on that dimension. Our approach is based on the hypothesis that, if a set of verbs apply to the same objects, they apply for similar reasons. We compile applications of action verbs to nouns naming objects in large text corpora, and use the resulting dataset to produce an embedding. This embedding represents each object as a vector in a low-dimensional space, where each dimension groups verbs that apply to similar objects.
%Further, we show that the dimensions learned are interpretable, and they group together verbs that would all typically occur during certain complex interactions with objects.
%
Our second goal is to understand the degree to which affordance knowledge underlies the mental representation of objects, as instantiated in SPoSE. We do this by showing that most dimensions of the SPoSE representation of an object can be predicted from its affordance embedding, in particular those that are categorical or functional.

%
%
%Our embedding approach  We start by identifying applications of action verbs to nouns naming objects, in large text corpora. We then u
%We do this for larger lists of objects and action verbs than previous studies (thousands in each case). 
%Combining the weights on each verb assigned by various dimensions yields a ranking over verbs for each concept. We show that this allows us to predict which verbs will be applicable to a given object, as captured in human judgments of affordance.
%Further, we show that the dimensions learned are interpretable, and they group together verbs that would all typically occur during certain complex interactions with objects.
%Finally, we show that they can be used to predict most dimensions of the SPoSE representation,  
%This suggests that affordance knowledge underlies much of the mental representation of objects, in particular semantic categorization.  

\section{Related Work}

The problem of determining, given an action and an object, whether the action can apply to the object was defined as {\em affordance mining} \cite{chao2015mining}. The authors proposed complementary methods for solving the affordance mining problem by predicting a plausibility score for each combination of object and action. 
%
%The best method used word co-occurrences in two ways: n-gram counts of verb-noun pairs, or similarity between verb and noun vectors in Latent Semantic Analysis~\cite{deerwester1990indexing} or Word2Vec~\cite{mikolov2013efficient} word embeddings. 
%
%For evaluation, they collected AMT judgements of plausibility (``is it possible to $<verb>$ a $<object>$'') for every combination of 91 objects and 957 action verbs. The authors found they could retrieve a small number of affordances for each item, but precision dropped quickly with a higher recall. 
%
Subsequent work \cite{rubinstein2015well,lucy2017distributional,forbes2019neural,utsumi2020exploring} predicted properties of objects in the norms above from word embeddings \cite{mikolov2013efficient,pennington2014glove}, albeit without a focus on affordances. %\cite{forbes2019neural} extracted 50 properties (some were affordances) from \cite{devereux2014centre}, for a set of 514 objects, to generate positive and negative examples for 25,700 combinations. They used this data to train a small neural network to predict these properties. The input to the network was either the product of the vectors for object and property, if using word embeddings \cite{mikolov2013efficient,pennington2014glove,levy2014dependency}, or the representation of a synthesized sentence combining them, if using contextualized embeddings \cite{peters2018deep,devlin2018bert}. They found that the latter outperformed the former for property prediction, but none allowed reliable affordance prediction.
In addition to object/action plausibility prediction, \citeA{ji2020functionality} addressed the problem of determining whether a object1/action/object2 relationship was plausible.
%They selected a set of 20 actions from \cite{chao2015mining} and combined them with the 70 most frequent objects in ConceptNet \cite{speer2012representing} into 1400 object/action pairs, which were then labelled as plausible or not; given rater disagreements, this yielded 330 positive pairs and 1070 negative ones. They then combined the positive pairs with other objects as "tails" (recipients of the action), yielding 3900 triplets. They reached F1 scores of 0.81 and 0.52 on the two problems, respectively. 
Other papers have focused on understanding the relevant visual features in objects that predict affordances \cite{myers2015affordance,sawatzky2017weakly,wang2020learning}. This has been combined with text in robotics literature, but usually focusing on a restricted set of objects and manipulation actions.
%The latter collected affordance judgments on AMT (``what can you do with $<object>$'') for 500 objects and harmonized them with WordNet synsets for 334 action verbs. 
For validation of the rankings of verb applicability predicted by our model, we will use the datasets from \cite{chao2015mining} and \cite{wang2020learning}, as they are the largest available human rated datasets. 
%
%In robotics research, affordance refer to relation between agent, action and the environment, under the constraints of motor and sensing capabilities of the agent \cite{lopes2007affordance}. Affordance modeling for robotics have been studied extensively, see recent surveys \cite{jamone2016affordances, zech2017computational, hassanin2018visual}. Due to the restriction in action possibilities and the complexity of real world scenarios, poor semantic generalization of affordance inference is observed in visual and speech-based robotic systems. Attempts to use semantic role labelling or text corpora, in tandem with visual features, have been proposed (e.g \cite{persiani2019unsupervised, chen2020enabling}. In general, though, this work focuses on a restricted set of objects and manipulation actions.
%
In computational linguistics, \citeA{resnik1993selection} introduced computational approaches to determining {\em selectional preference}, the degree to which a particular semantic class tends to be used as an argument to a given predicate. Several methods have been proposed to do this, leveraging some grouping of verbs and objects into classes (e.g. WordNet in \cite{resnik1996selectional}, or co-occurrence statistics of words in a corpus \cite{Erk2007,Pado2007,Seaghdha2010,VanDeCruys2014,zhang2020multiplex}. 
All of these methods could be used to score verbs by how applicable they are to a given noun, the ancillary task we use to make sure our embedding carries the relevant information. % That said, they would require substantial implementation effort, even where the code is publicly available.
%This said, making that prediction is not our main goal. 
 Our proposed embedding space is a latent variable model for verb-noun applications. While this is also the case for these papers, they would require extensive modification to add sparsity assumptions -- important for interpretability -- and to produce verb rankings.

\section{Methods}

\subsection{Objects and Actions considered}

We used the list of 1854 object concepts introduced in \cite{hebart2019things} and for which SPoSE embeddings are available. This list sampled  from concrete, picturable, and nameable nouns in American English, and was further expanded by an AMT study to elicit category\footnote{Main categories: food, animal, clothing, tool, drink, vehicle, fruit, vegetable, body part, toy, container, bird, furniture, sports equipment, musical instrument, dessert, part of car, weapon, plant, insect, kitchen tool, office supply, clothing accessory, kitchen appliance, home decor, medical equipment, and electronic device.} memberships. As we are not doing sense disambiguation for each noun that names an object, we will use "noun" or "object" interchangeably.
 We created our own verb list by having three annotators go through all verb categories on VerbNet~\cite{schuler2005verbnet}, and selecting those that included verbs that corresponded to an action\footnote{Those VerbNet categories contained $\sim 10-50$ verbs sharing thematic roles and selectional preferences (e.g. fill-9.8, amalgamate-22.2, manner-speaking-37.3, build-26.1, remove-10.1, cooking-45.3, create-26.4, destroy-44, mix-22.1, vehicle-51.4.1, dress-41.1.1).} performed by a human on an object. We kept only those categories where all annotators agreed, and all verbs in each category. The resulting list has 2541 verbs.

\subsection{Extraction of Verb Applications to Nouns from Text}
\label{sec:PPMI}

We used the UKWaC and Wackypedia corpora~\cite{ferraresi2008introducing}, with approximately, 2B and 1B tokens, and 88M and 43M sentences, respectively. The former is the result of a crawl of British web pages, while the latter is a subset of Wikipedia. Both have been cleaned and have clearly demarcated sentences, which is ideal for dependency parsing. We replaced all common bigrams in \cite{brysbaert2014concreteness} by a single token. 

We identified all sentences containing both verbs and nouns in our list, and we used Stanza
%\cite{qi2020stanza}
to produce dependency parses for them. We extracted all the noun-verb pairs in which the verb was a syntactic head of a noun having \texttt{obj} (object) or \texttt{nsubj:pass} (passive nominal subject) dependency relations. We compiled raw counts of how often each verb was used on each noun within a sentence, producing a count matrix $M$. Note that this is different from normal co-occurrence counts; those would register a count whenever verb and noun were both present within a short window (e.g. up to 5 words away from each other), {\em regardless} of whether the verb applied to the noun, or they were simply in the same sentence. 
%Out of the 1854 nouns considered, there were 1755 with at least one associated action.
Note also that the counts pertain to every possible meaning of the noun.% given that no word sense disambiguation was performed.

Finally, we converted the matrix $M$ into a Positive Pointwise Mutual Information (PPMI~\cite{turney2010frequency}) matrix $P$ where, for each object $i$ and verb $k$:
\begin{equation}
P(i,k) := \max\left(\log \frac{\mathbb{P}(M_{ik})}{\mathbb{P}(M_{i*}) \cdot \mathbb{P}(M_{*k})},0\right), 
\label{eqt:ppmi}
\end{equation}
$\mathbb{P}(M_{i*})$ and $\mathbb{P}(M_{*k})$ are marginal probabilities of $i$ and $k$. %$P$ can be viewed as a pair-pattern matrix~\cite{lin2001dirt}, where the PPMI helps in separating frequency from informativess of the co-occurrence of nouns and verbs \cite{turney2010frequency, turney2003measuring}. However, PPMI is also biased and may be large for rare co-occurrences, e.g. for $(o_i, v_k)$ that co-occur only once in $M$. This is addressed in the process described in the next section.

\begin{table}[htb]
\begin{center}

\caption{Top 5 verbs in selected affordance dimensions.}
%\caption{Verb assignment for each affordance embedding dimension} 
\label{table:assignment_dimension_selection}
% \vskip 0.12in
%\footnotesize
\scriptsize 
\begin{tabular}{rl}
Dimension & Top 5 verbs in each affordance dimension\\
\midrule
%1 & invent, intrdc, mnfctr, dvlp, design\\
1 & invent, introduce, manufacture, develop\\ %design\\
2 & blanch, boil, steam, drain, cook\\
3 & spot, observe, sight, hunt, watch\\
4 & park, drive, hire, crash, rent\\
5 & wield, grab, carry, hold, hand\\
6 & squirt, formulate, dilute, smear, dissolve\\
7 & capsize, moor, sail, beach, raft\\
8 & grass, uproot, mulch, smother, clothe\\
9 & wear, don, unbutton, match, button\\
10 & coil, splice, braid, sever, thread\\
11 & rouge, twinkle, flinch, twitch, sneer\\
12 & mewl, breast, coo, breastfeed, swaddle\\
13 & empty, fill, clean, clutter, line\\
14 & tiptoe, totter, leer, yowl, mosey\\
15 & serve, eat, cook, prepare, order\\
16 & drink, sip, sup, swig, quaff\\
17 & determine, compute, plot, ascertain\\
% underline\\
18 & pasture, herd, slaughter, milk, tether\\
19 & moo, pomade, gel, tweeze, primp\\
20 & weave, drape, embroider, knit, sew\\
21 & lob, hurl, fire, throw, explode\\
22 & wet, moisten, rinse, soak, reuse\\
23 & fleck, scallop, strew, emanate, pluck\\
24 & sound, hear, play, blare, amplify\\
25 & bare, swathe, waver, thump, tattoo\\
26 & steal, recover, retrieve, discover, hide\\
27 & freckle, moisturize, spritz, dehair, deflesh\\
28 & close, open, shut, padlock, unlatch\\
29 & sprinkle, mix, add, stir, blend\\
%30 & meow, fissure, furrow, burble, decoct\\
31 & manufacture, buy, purchase, sell, design\\
32 & dodder, skedaddle, snicker, roust, sober\\
33 & extinguish, light, kindle, rekindle, flare\\
34 & strangulate, fumble, glove, punt, bunt\\
35 & unscrew, screw, slacken, disengage, tighten\\
36 & declaw, leash, worm, feud, groom\\
37 & hunt, kill, cull, exterminate, chase\\
38 & unfasten, tighten, fasten, undo, loosen\\
39 & dodder, skedaddle, snicker, roust, sober\\
40 & deice, whir, flit, swagger, quiver\\
%41 & lignify, distemper, burl, stucco, paper\\
%42 & birch, remonstrate, billet, room, feminize\\
43 & cloister, remarry, ostracize, unionize, intermarry\\
%44 & nationalize, telephone, reorganize\\
%& liquidate, federate\\
45 & gabble, cluck, bridle, loll, lisp\\
%46 & energize, implant, cool\\
%& interconnect, couple\\
47 & winnow, mill, parboil, grind, reap\\
%48 & quarry, vein, petrify, deforest, silicify\\
49 & grill, baste, barbecue, marinate, brown\\
50 & sharpen, blunt, wield, plunge, thrust\\
51 & thicken, spoon, reheat, stir, simmer\\
52 & sprain, hyperextend, flex, fracture, injure\\
%53 & honeycomb, etch, weld, lacquer, emboss\\
54 & eradicate, deter, swat, combat, discourage\\
%55 & lour, eddy, scud, glower, taxi\\
%56 & ossify, deflesh, calcify, masticate, jumble\\
57 & cultivate, grow, plant, prune, propagate\\
58 & pilot, board, rearm, crew, station\\
%59 & ulcerate, distend, decompress\\
%& sicken, repopulate\\
%60 & motorcycle, zigzag, crisscross, roil, span\\
61 & install, connect, disconnect, activate, operate\\
62 & erect, carve, flank, adorn, construct\\
63 & fish, catch, destress, whiff, degut\\
64 & bake, leaven, ice, eat, serve\\
65 & block, clog, dam, choke, flood\\
66 & fit, mount, position, incorporate, attach\\
67 & slice, peel, chop, dice, grate\\
68 & unload, wheel, lug, load, transport\\
%69 & scuffle, tramp, shovel, whoosh, heap\\
70 & munch, scoff, eat, gobble, nibble\\
\midrule
\end{tabular}
\end{center}
\end{table}

\subsection{Object embedding in a verb usage space}
\label{Sec:NMF}

\paragraph{Object embedding via matrix factorization}

Our embedding is based on a factorization of the PPMI matrix $P$ ($m$ objects by $n$ verbs) into the product of matrices $O$ ($m$ objects by $d$ dimensions) and $V$ ($n$ verbs by $d$ dimensions), yielding $\widetilde{P} := OV^T \approx P$. $O$ is the object embedding in $d$-dimensional space, and $V$ is the weighting of each verb in each dimension.  Each column $V_{:,k}$ of matrix $V$ contains a pattern of verb usage for {\em dimension} $k$, which captures verb co-occurrence across all objects. Intuitively, if two verbs occur often with the same objects, they will both have high loadings on one of the $d$-dimensions; conversely, the objects they occur with will share high loadings on that dimension. The top-5 verb patterns for most of the 70 dimensions are shown in Table~\ref{table:assignment_dimension_selection}.

The idea of factoring a count matrix (or a transformation of it) dates back to Latent Semantic Analysis \cite{landauer1997solution}, and was investigated by many others \cite{turney2010frequency}.
%Given that PPMI is biased towards rare pairs of noun/verb, the matrix $P$ is not necessarily very sparse. 
If factorized into a product of two low-rank matrices, the structure of the matrix can be approximated while excluding noise or rare events. %\cite{bullinaria2012extracting}. 
%
%\paragraph{Optimization problem}
%
Given that the PPMI matrix $P$ is positive, the matrices $O$ and $V$ are as well. We obtain them through a non-negative matrix factorization (NMF) problem
\begin{equation}
O^*, V^* = \underset{O, V}{\text{argmin}} \Vert P - OV^T \Vert_F^2 + \beta \mathcal{R}(O, V),
\label{eqt:nmf}
\end{equation}
which can be solved through an iterative minimization procedure. For the regularization $\mathcal{R}(O, V)$, we chose the sparsity control $\mathcal{R}(O, V) \equiv \sum_{ij} O_{ij} + \sum_{ij} V_{ij}$. 
We used the NNDSVD initialization, 
%~\cite{boutsidis2008svd}, 
a SVD-based initialization which favours sparsity on $O$ and $V$ and approximation error reduction.
We found that the optimal dimensionality and sparsity were $d=70$ and $\beta=0.3$, respectively, using the two-dimensional hold-out cross validation procedure described in the Appendix.
%\label{appendix:hyperparameter}
%~\cite{kanagal2010rank}  
This procedure removes entire blocks of the matrix at a time, and reconstructs them using a decomposition of the rest of the matrix, using a range of dimensionality and sparsity settings.
%These values were found using two-dimensional hold-out cross validation.
%the two-dimensional hold-out cross validation~\cite{kanagal2010rank}, due to its scalability and natural fit to the multiplicative update algorithm for solving (\ref{eqt:nmf}). 
%
%Specifically, denoting $M_t$ and $M_v$ to be the mask matrices representing the held-in and held-out entries, we optimize for
%\begin{equation}
%    O^*, V^* = \underset{O, V}{\text{argmin}} \Vert M_t \odot (P - OV^T) \Vert_F^2 + \beta \mathcal{R}(O, V)
%\end{equation}
%and obtain the reconstruction 
%
%error $E = \Vert M_v \odot (P - O^*(V^*)^T) \Vert_F^2 + \beta \mathcal{R}(O^*, V^*)$. %For more details, please refer to Appendix~\ref{appendix:hyperparameter}. 
%
%The optimization problem (\ref{eqt:nmf}) is NP-hard and all state-of-the-art algorithms may converge only to a local minimum~\cite{gillis2014and}; choosing a proper initialization of $O$ and $V$ is crucial. 
% 
%

\paragraph{Estimating the verb usage pattern for each object}
\label{sec:embedding_transformation}

Deriving a similar pattern for each {\em object} $i$, given its embedding vector $O_{i,:} = [o_{i_1}, o_{i_2}, \ldots o_{i_d}]$, requires combining these patterns based on the weights given to each dimension. This requires computing the cosine similarity between each embedding dimension $O_{:,h}$ and the PPMI values $\tilde{P}_{:,k}$ for each verb $k$ in the approximated PPMI matrix $\tilde{P} = O V^T$, which is
\begin{equation}
S(O_{:,h}, \tilde{P}_{:,k}) = \frac{ O_{:,h} \cdot \tilde{P}_{:,k} }{\Vert O_{:,h} \Vert_2 \Vert \tilde{P}_{:,k} \Vert_2}.
\label{eqt:cosine_score}
\end{equation}

Given the embedding vector for object $i$, $O_{i,:} = [o_{i_1}, o_{i_2}, \ldots o_{i_d}]$, we compute the pattern of verb usage for the object as $O_{i,:} S$.
Thus, this is a weighted sum of the similarity between every $O_{:,h}$ and $\tilde{P}_{:,k}$. We will refer to the ordering of verbs by this score as the  {\em verb ranking} for object $i$.

\section{Experiments and Results}

\subsection{Prediction of affordance plausibility}

\paragraph{Affordance ranking task}

The first quantitative evaluation of our embedding focuses on the ranking of verbs as possible affordances for each object. We will use the Affordance Area Under The Curve (AAUC) relative to datasets that provide, for each object, a set of verbs known (or likely) to be affordances. Intuitively, the verb ranking for object $i$ is good if it places these verbs close to the top of the ranking, yielding an AAUC of 1. Conversely, a random verb ranking would have an AAUC of 0.5, on average. This is a conservative measure, given that a perfect ranking would still penalize every true affordance not at the top. Hence, this is useful as a {\em relative} measure for comparing between our and competing approaches for producing rankings. More formally, given the $K$ ground truth verb affordances $\{g_k\}_{k=1}^K$ of object $i$, and its verb ranking $\{ v_i \}_{i=1}^n$, we denote $\ell_k$ to be the index such that $v_{\ell_k} = g_k \ \forall k$. We then define AUCC for object $i$ as $\text{AUCC} = \tfrac{1}{K} \sum\nolimits_{k=1}^K \left(1 - \tfrac{\ell_k}{n}\right)$.

\paragraph{Datasets}
\label{sec:experiments.prediction.datasets}

We use the two largest publicly available object affordance datasets as ground truth. In the first dataset, WTAction~\cite{wang2020learning}, objects are associated with the top 5 actions label provided by human annotators in response to ``What can you do with this object?''. Out of 1,046 objects and 445 actions in this dataset, there are 971 objects and 433 verbs that overlap with those in our lists ($\sim$ 3.12 action labels per object) . The second dataset, MSCOCO~\cite{chao2015mining}, scores every candidate action for an object ranging from 5.0 (``definitely an affordance'') to 1.0 (``definitely not an affordance''). We consider only a 5.0 score as being an affordance.
%, whereas \cite{chao2015mining} used both 4.0 and 5.0.
Out of 91 objects and 567 actions, 78 objects and 558 verbs overlap with ours ($\sim$ 34 action labels per object). 

\paragraph{Baseline methods}

We compared the ranking of verbs produced by our algorithm with an alternative proposed in \cite{chao2015mining}: ranking by the cosine similarity between word embedding vectors for each noun and those for all possible verbs in the dataset. We considered several off-the-shelf embedding alternatives, namely Word2Vec (\cite{mikolov2013efficient}, 6B token corpus), GloVe (\cite{pennington2014glove}, 6B and 840B token corpora, Dependency-Based Word Embedding (DBWE, \cite{levy2014dependency}, 6B corpus), and Non-negative Sparse Embedding (NNSE, \cite{Murphy2012b}, 16B corpus). 
%Finally, we included the other two methods in \cite{chao2015mining}, LSA (\cite{deerwester1990indexing}, trained on our corpus), and ranking by frequency of verb/noun pair in Google N-grams (\cite{lin2012syntactic}).
The embeddings are 300-D in all cases, except for NNSE (1000-D, similar results for 2500-D). Finally, we also ranked the verbs by their values in the row of the PPMI matrix P for each probed object, to see how much our method of embedding through a low-rank approximation allowed the extraction of additional information.

\cob
%We contrasted Word2Vec and GloVe because they are based on two different embedding approaches (negative sampling and decomposition of a word co-occurrence matrix), developed on corpora twice as large as ours. We contrasted 6B and 840B versions of GloVe to see the effect of increasing dataset size, and 2B to show the effect of our corpus. In these methods, the co-occurrence considered is simply proximity within a window of a few tokens, rather than application of the verb to the noun. \cor We included DBWE because it uses dependency parse information (albeit to define the word co-occurrence window, rather than select verb applications to nouns as we do). We included NNSE because it is based on a sparse, non-negative factorization of a co-occurrence matrix of features derived from words combined with specific dependency relations. \cob Finally, we ranked the verbs by their values in the row of the PPMI matrix $P$ corresponding to each probed object, to see the effect of using a low-rank approximation in extracting information.

\begin{table}[H]
\caption{AAUC of verb rankings by each method.}
\label{tab:results-20200928}
\scriptsize
% \footnotesize
\begin{tabular}{crrrrrrr}
Dataset & \multicolumn{7}{c}{Method} \\
\midrule
    & DBWE & NNSE & W2V & GV & G840 & LSA & Ours \\
WTA & 0.60 & 0.65 & 0.70 & 0.75 & 0.80 & 0.81 & 0.88 \\
MSC & 0.56 & 0.58 & 0.59 & 0.65 & 0.68 & 0.63 & 0.77 
\end{tabular}
\end{table}

\paragraph{Results}

For each dataset, we reduced our embeddings $O$ and $V$ according to the sets of objects and verbs available. We then obtained the verb ranking for each object, as described in the {\bf Methods} section, as well as rankings predicted with the different baseline methods in the previous section. Table~\ref{tab:results-20200928} shows average AAUC results 
%(in terms of the mean $\mu$ and the standard error $S_E$ of AAUCs) 
obtained with these verb rankings on the two datasets. Our ranking is better than those of all the baseline methods, as well as PPMI (0.77, 0.61), as determined from paired two-sided $t$-tests, in both WTAction %($p$-values of $7.36\mathrm{e}{-23}$, $4.19\mathrm{e}{-14}$, $6.82\mathrm{e}{-174}$, $1.34\mathrm{e}{-53}$, $1.44\mathrm{e}{-89}$, $5.21\mathrm{e}{-47}$, $8.70\mathrm{e}{-25}$ and $8.14\mathrm{e}{-39}$)
and MSCOCO
(all $p$-values $\ll 0.01$).
%, and significant correcting for multiple comparisons)
%($p$-values of $1.54\mathrm{e}{-23}$, $3.37\mathrm{e}{-21}$, $9.32\mathrm{e}{-36}$, $2.15\mathrm{e}{-27}$, $2.82\mathrm{e}{-30}$, $9,64\mathrm{e}{-20}$, $7.93\mathrm{e}{-17}$ and $2.13\mathrm{e}{-29}$).
%
%The overall distributions of AAUCs for the two datasets are shown in Appendix~\ref{appendix:AAUC}. Our procedure yields AAUC closer to 1.0 for many more items than the other methods. This suggests that the co-occurrence of verb and noun within a window of a few tokens, the basis of the word embeddings that we compare against, carries some information about affordances, but also includes other relationships and noise (e.g. if the verb is in one clause and the noun in another). Results are better in the embedding trained in a corpus 280X larger, but still statistically worse than those of our procedure. Ranking based on the PPMI matrix $P$ performs at the level of the 6B token embeddings. This suggests that our procedure is effective at removing extraneous information in the matrix $P$.
%
The following figure contrasts the AAUC distribution across objects for our method with those obtained with the top 4 embeddings and PPMI, on the WTaction and MSCOCO datasets, respectively.

\begin{figure}[htb]
    \begin{center}
    \includegraphics[width=1\linewidth]{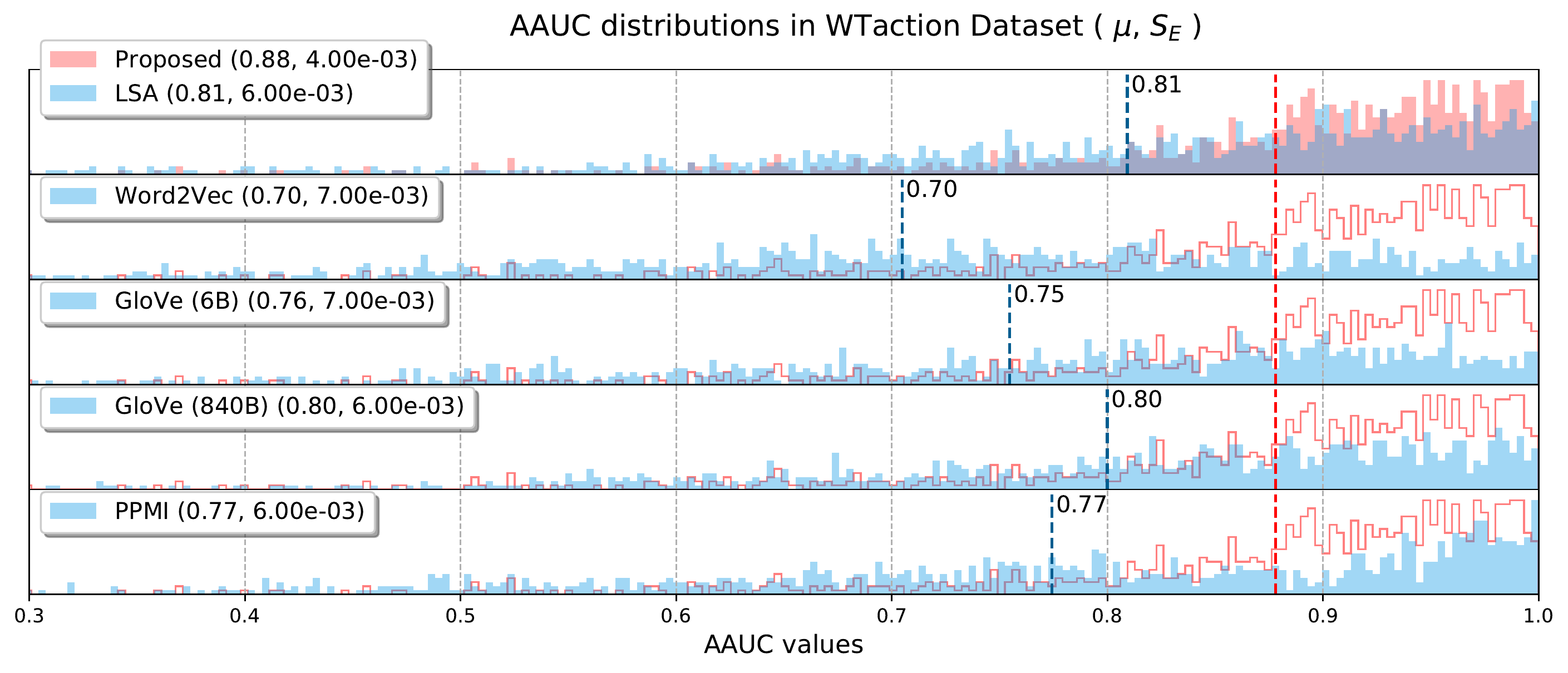}
    \end{center}
    %\caption{AAUC Distribution on WTaction Dataset}
%    \label{fig:Aria-AAUC}
%\end{figure}

%\begin{figure}[htb]
    \begin{center}
    \includegraphics[width=1\linewidth]{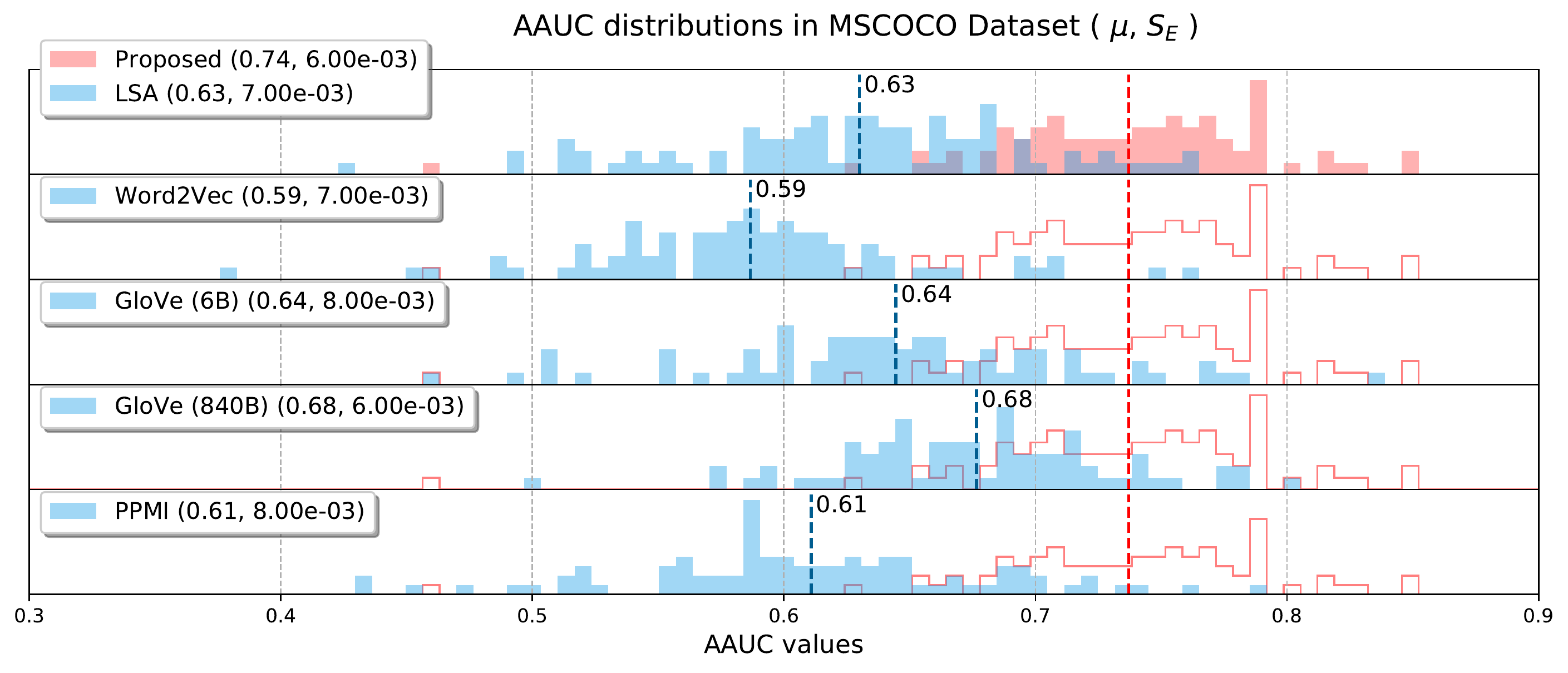}
    \end{center}
    %\caption{AAUC Distribution on MSCOCO Dataset}
%    \label{fig:Mihalcea-AAUC}
    \caption{AAUC distribution on WTaction (top) and MSCOCO (bottom) datasets using our method, against the 4 top embeddings and PPMI.}
    \label{fig:AAUC}
\end{figure}

\subsection{Prediction of SPoSE object representations}
\label{sec:SPoSE}

\paragraph{The SPoSE representation and dataset}

The dimensions in the SPoSE representation~\cite{hebart2020revealing} are interpretable, in that human subjects coherently label what those dimensions are ``about'', from the categorical (e.g. animate, building) to the functional (e.g. can tie, can contain, flammable) or structural (e.g. made of metal or wood, has inner structure).
%Furthermore, subjects could also predict what dimension values new objects would have, based on knowing the dimension value for a few other objects.
The SPoSE vectors for objects are derived from behaviour in a ``which of a random triplet of objects is the odd one out'' task. The authors propose a hypothesis for why there is enough information in this data to allow this: when given any two objects to consider, subjects mentally sample the contexts where they might be found or used. The resulting dimensions reflect the aspects of the mental representation of an object that come up in that sampling process. The question we want to answer is, then, which of these dimensions reflect affordance or interaction information. We used the 49-D SPoSE embedding published with \cite{hebart2020revealing}. We excluded objects named by nouns that had no verb co-occurrences in our dataset and, conversely, verbs that had no interaction with any objects. We averaged the vectors for objects named by the same polysemous noun (e.g. ``bat''). The resulting dataset had 1755 objects/nouns, and 2462 verbs.  

%\begin{figure}[htb]
%    \begin{center}
%    \includegraphics[width=1\linewidth]{Figures/SPoSE_approximation_pattern-20200928-v2-10.pdf}
%    \end{center}
%    \caption{Prediction of first 10 SPoSE embeddings from affordance embeddings (top) versus actual SPoSE embeddings (middle). Objects are grouped by semantic category (those with $\ge 15$ objects). The absolute residuals of the prediction are also shown (bottom). Color range is fixed across plots to show the magnitude of residuals. For full diagram with all dimensions, see Figure~\ref{fig:SPoSE_approximation_pattern_full} in  Appendix~\ref{appendix:SPoSE_pattern}.}
%    \label{fig:SPoSE_approximation_pattern}
%\end{figure}

\begin{figure}[htb]
        \centering
        \includegraphics[width=1\linewidth]{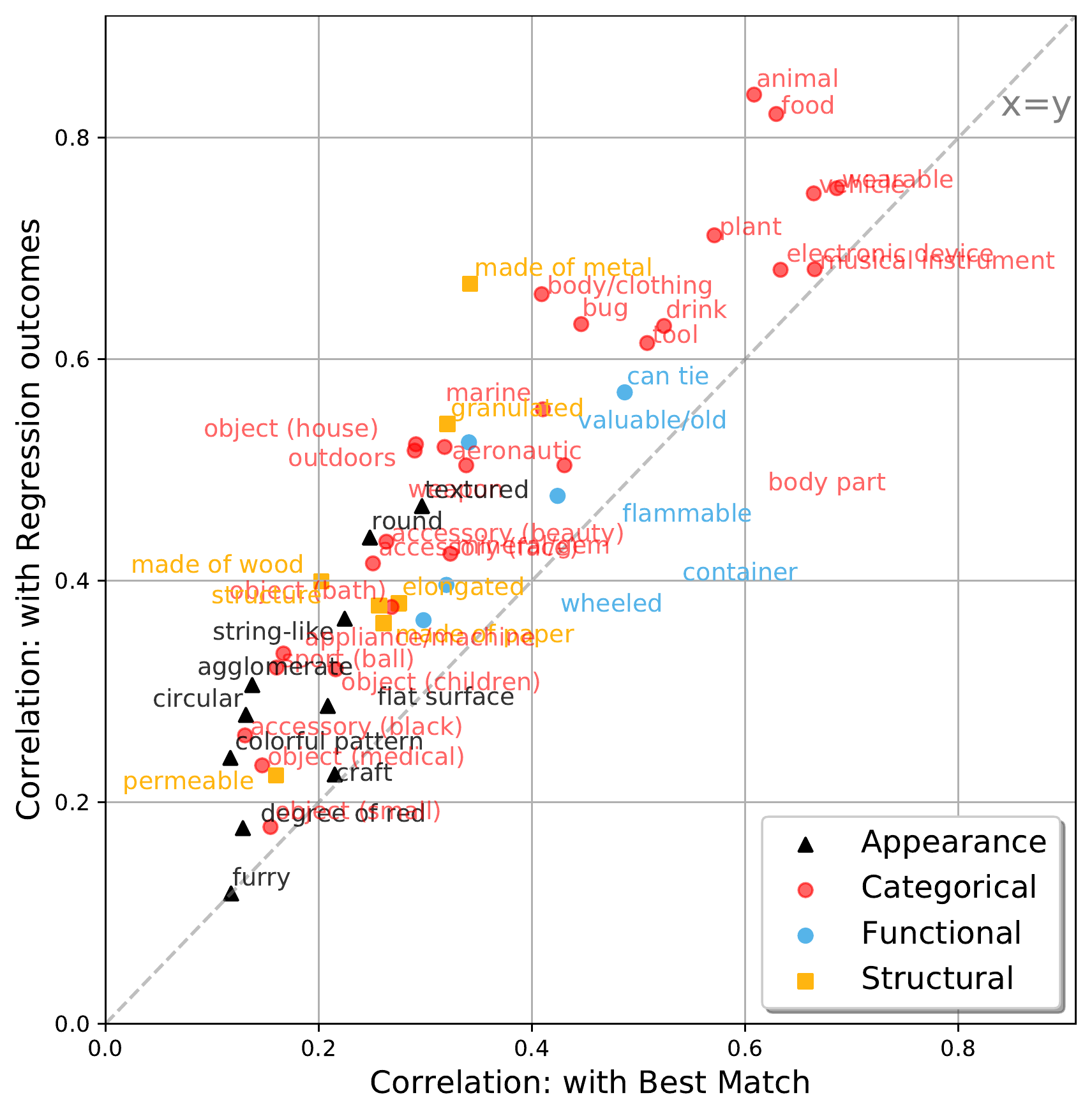}  
        \label{fig:SPoSE_cosine_plot}
        \vspace{-1em}
\caption{For each SPoSE dimension, correlation with the best matching affordance dimension (x-axis) and with the cross-validated prediction of the regression model for it (y-axis).}
\label{fig:SPoSE_approximation_similarity}
\end{figure}

\paragraph{Relationship between SPoSE and affordance dimensions}

We first considered the question of whether affordance dimensions correspond directly to SPoSE dimensions, by looking for the highest correlated match.
%As shown in Figure~\ref{fig:SPoSE_approximation_similarity}(a),
Many of the 49 SPoSE dimensions are similar to at least one of the 70 affordance dimensions, with the distribution of correlation of the best match shown in the x-axis of Figure~\ref{fig:SPoSE_approximation_similarity}. Then,
%\paragraph{Prediction of SPoSE dimensions from affordance embedding}
in order to determine which SPoSE dimensions of an object could be explained in terms of affordance dimensions, we predicted their value from the affordance embedding of the object. Denoting the SPoSE vectors for $m$ objects as a $m \times 49$ matrix Y, we solved this Lasso regression problem for each column $Y_{:,i}$
\begin{equation}
    w_i^* = \operatorname*{argmin}_{w \in \mathbb{R}^{d}, w \geq 0}\tfrac{1}{2m} \left\Vert Y_{:,i} - O w \right\Vert^2_2 + \lambda \Vert w \Vert_1, \quad i = 1,\ldots 49,
\label{eqt:regression}
\end{equation}
where $\lambda$ was chosen based on a 2-Fold cross-validation, with $\lambda$ in $[10\mathrm{e}^{-7}, 10\mathrm{e}^{3}]$ with log-scale spacing. Since both $Y_{:,i}$ and our embedding $O$ represent object features by positive values, we restricted $w\geq0$. Intuitively, this means that we try to explain every SPoSE dimension by combination of the {\em presence} of certain affordance dimensions, not by trading them off.

\begin{figure}[!h]
    \includegraphics[width=0.99\linewidth]{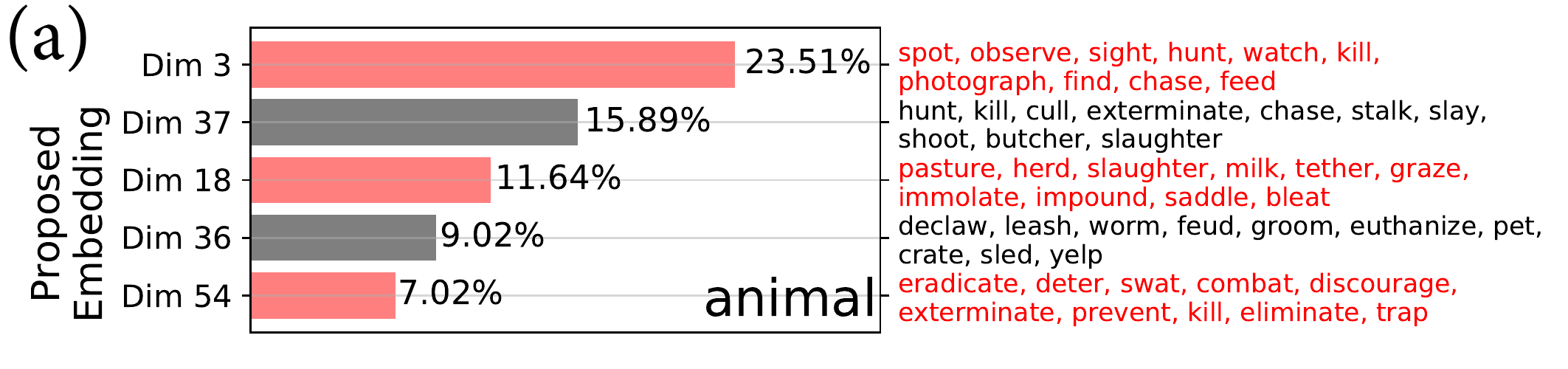}
    \includegraphics[width=0.99\linewidth]{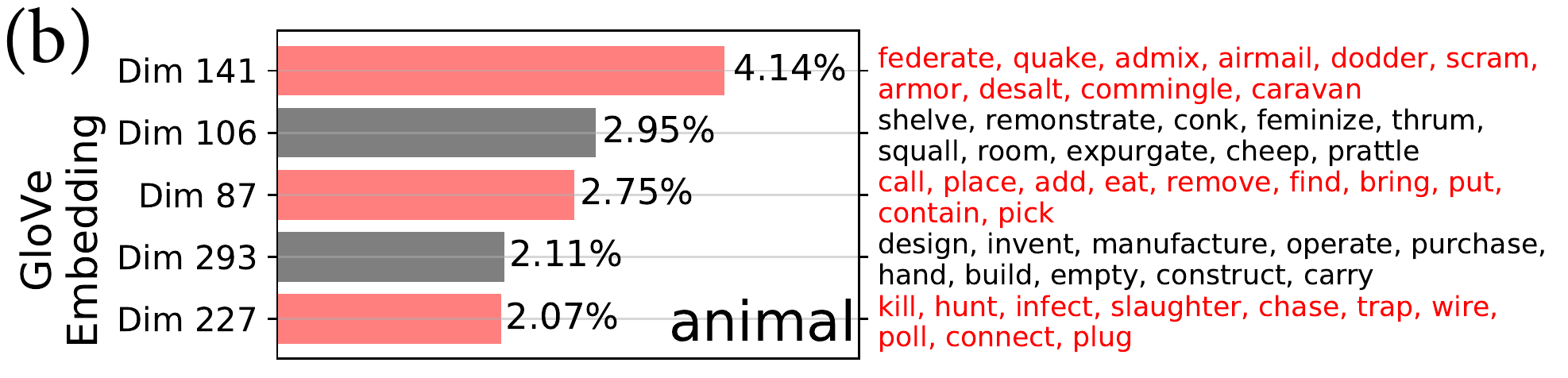}
  \caption{ Top 10 verbs in the 5 most important affordance dimensions (proposed affordance embedding versus GloVe 840B) for predicting the ``animal'' SPoSE feature. \label{fig:SPoSE_component_compare}}
\end{figure}

Overall, the cross-validated predictions of this regression model are more similar to SPoSE dimensions than any individual affordance dimension, as shown in the y-axis of Figure~\ref{fig:SPoSE_approximation_similarity}. The best predicted dimensions are categorical, e.g. ``animal'', ``plant'', or ``tool'', or functional, e.g. ``can tie'' or ``flammable''. Structural dimensions are also predictable, e.g. ``made of metal'', ``made of wood'', or ``paper'', but less so for appearance-related dimensions, e.g. ``colorful pattern'', ``craft'', or ``degree of red''.
%
%However, when we consider the cross-validated regression models to predict SPoSE dimensions from affordance dimensions, {\em every} model places non-zero weight on several of the latter. Furthermore, those predictions are much 
%
What can explain this pattern of predictability? Most SPoSE dimensions can be expressed as a linear combination of affordance dimensions, where both the dimensions and regression weights are {\em non-negative}. This leads to a sparse regression model -- since dependent variables cannot be subtracted to improve the fit -- where, on average, 5 affordance dimensions have 80\% of the regression weight. Each affordance dimension, in turn, corresponds to a ranking over verbs. Figure~\ref{fig:SPoSE_component_compare}a shows the top 10 verbs in the 5 most important affordance dimensions for predicting the ``animal'' SPoSE dimension. As each affordance dimension loads on verbs that correspond to broad modes of interaction (e.g.  observation,  killing, husbandry), the model is both predictive and interpretable. Whereas we could use dense embeddings to predict SPoSE dimensions, they do not work as well (in either accuracy or interpretability, see Figure~\ref{fig:SPoSE_component_compare}b for GloVe).
%\begin{figure}[!h]
%    \includegraphics[width=0.51\linewidth]{Figures/SPoSE_component_animal-proposed-20200928.pdf}
%    \includegraphics[width=0.48\linewidth]{Figures/SPoSE_component_animal_Glove-20200928.pdf}
%  \caption{ Top 10 verbs in the 5 most important affordance dimensions (proposed affordance embedding versus GloVe 840B) for predicting the ``animal'' SPoSE feature. \label{fig:SPoSE_component_compare}}
%\end{figure}
%
%
For example, if we consider the top 5 verbs from affordance dimensions used in predicting each SPoSE dimension, we see that ``tool'' has ``sharpen, blunt, wield, plunge, thrust'' (D50); 
%``animal'' has ``spot, observe, sight, hunt, watch'' (D3), ``hunt, kill, cull, exterminate, chase'' (D37), or ``pasture, herd, slaughter, milk, tether'' (D18);
``food'' has ``serve, eat, cook, prepare, order'' (D15), or ``bake, leaven, ice, eat, serve'' (D64); ``plant'' shares D2 with ``food'', but also has ``cultivate, grow, plant, prune, propagate'' (D57). 

%The full list of affordance dimensions most relevant for predicting each SPoSE dimension is in Appendix~\ref{appendix:SPoSE_components} (Figure~\ref{fig:contributionA}--\ref{fig:contributionF}).
%
These results suggest that SPoSE dimensions are predictable {\em insofar} as they can be expressed as combinations of modes of interaction with objects. As described in {\bf Methods} section, we can combine affordance dimension verb rankings into a verb ranking for each SPoSE dimension. We replaced the embedding $O$ in (\ref{eqt:cosine_score}) with the SPoSE prediction $\widetilde{Y}$ and we ranked the verbs for dimension $h$ according to $S(\tilde{Y}_{:,h}, \tilde{P}_k)$. Table~\ref{table:assignment} shows, for every SPoSE dimension, ranked by predictability, the top 10 verbs in its ranking. 
%
%Testing against a null hypothesis of $0$ correlation, the p-values obtained range from 0.0 to the maximum of 7.61\mathrm{e}{-7} for the SPoSE dimension ``furry'', indicating the statistical significance of correlation with our regression outcomes.
%
This table suggests that
%, first and foremost, 
highly predictable categorical dimensions correspond to very clean affordances. The same is true for functional dimensions, e.g. ``can tie'' or ``container'' or ``flammable''; even though they are not ``classic'' categories, subjects group items belonging to them based on their being suitable for a purpose (e.g. ``fasten'', ``fill'', or ``burn''). Why would this hold for structural dimensions? One possibility is if objects having that dimension overlap substantially with a known category (e.g. ``made of metal'' and ``tool''). Another is that the structure drives manual or mechanical affordance (e.g. ``elongated'' or ``granulated''). Finally, what are the affordances for appearance dimensions that can be predicted? Primarily, actions on items in categories that share that appearance, e.g. ``textured'' is shared by fabric items, ``round'' is shared by many fruits or vegetables. Prediction is worse when the items sharing the dimension come from many different semantic categories.
%(\cite{hebart2019things} lists the pictures of items shown to subjects).

%\begin{comment}
%%%%%%%%%%%%%%%%%%%%%%%%%%%
% previous table
%%%%%%%%%%%%%%%%%%%%%%%%%%%
\begin{table*}[htb]
%\footnotesize
\caption{Affordance assignment for a selection of SPoSE dimensions mentioned in the text, ordered by how well they can be predicted from the affordance embedding. The names of SPoSE dimensions are simplified.} 
    %The full table, $p$-values and descriptions for each dimension are in Appendices~\ref{appendix:affordance_assignment} and ~\ref{appendix:SPoSE_description}, respectively.}
   \label{table:assignment}
\begin{center}
\scriptsize % scriptsize
\begin{tabular}{rlll}
Correlation & SPoSE dimension & Type & Affordances (Top Ten Ranked Verbs) \\
\midrule
0.84 & animal & categorical & kill, spot, hunt, observe, chase, feed, slaughter, sight, trap, find \\
0.82 & food & categorical & serve, eat, cook, prepare, taste, consume, add, mix, stir, order \\
0.75 & wearable & categorical & wear, don, match, knit, sew, fasten, rip, embroider, tear, model \\
0.71 & plant & categorical & grow, cultivate, plant, add, eat, chop, gather, cut, dry, prune \\
0.67 & made of metal & structural & fit, invent, manufacture, incorporate, design, position, attach, utilize, carry, install \\
0.61 & tool & categorical & wield, grab, hold, carry, sharpen, swing, hand, pick, clutch, throw \\
0.57 & can tie & functional & fasten, tighten, unfasten, undo, attach, thread, tie, secure, loosen, loose \\
0.54 & granulated & structural & contain, mix, scatter, add, gather, remove, sprinkle, dry, deposit, shovel \\
0.48 & flammable & functional & light, extinguish, ignite, throw, carry, flash, kindle, place, manufacture, douse \\
0.47 & textured & appearance & remove, place, hang, tear, stain, spread, weave, clean, drape, wrap \\
0.44 & round & appearance & grow, cultivate, pick, add, slice, place, eat, chop, throw, plant \\
0.40 & made of wood & structural & place, remove, carry, incorporate, design, contain, bring, construct, manufacture, find \\
0.40 & container & functional & empty, fill, carry, place, clean, load, bring, dump, unload, leave \\
0.38 & elongated & structural & grab, carry, wield, hold, pick, place, throw, hand, bring, drop \\
0.24 & colorful pattern & appearance & manufacture, buy, design, place, remove, sell, invent, purchase, contain, bring \\
0.23 & craft & appearance & place, bring, remove, design, hang, call, buy, put, pull, manufacture \\
0.22 & permeable & structural & fit, incorporate, remove, place, design, manufacture, install, position, clean, attach \\
0.18 & degree of red & appearance & place, call, add, contain, remove, find, buy, bring, introduce, sell \\
\end{tabular}
\end{center}
\end{table*}

\section{Conclusions}
\label{sec:discussion}

In this paper, we introduced an approach to embed objects in a space where every dimension corresponds to a pattern of verb applicability to those objects. We view such a pattern as a very broad extension of the classical notion of "affordance", obtained by considering verbs that go well beyond concrete motor actions, and objects that encompass many different categories beyond tools or household objects.
We showed that this embedding can be learned from a text corpus and used to rank verbs by how applicable they would be to a given object. 
%We evaluated this prediction against two separate human judgment ground truth datasets, and verified that our method outperforms general-purpose word embeddings trained on much larger text corpora. \cor This gave us with confidence that the patterns of applicability we identified captured information relevant to human judgments of affordance.\cob
%
We used our embedding to predict SPoSE dimensions for objects.
%SPoSE is an embedding that was derived from human behavioural judgements of object similarity, and has interpretable dimensions. 
%We used the resulting prediction models to probe the relationship between \cor the dimensions of our embedding and the various types of SPoSE dimensions. 
This allowed us to conclude that our "affordance" embedding knowledge predicts 1) category information, 2) purpose, and 3) some structural aspects of the object. SPoSE dimensions to do with visual appearance were poorly predicted.
%Our embedding is thus sufficient for predicting most SPoSE dimensions.
To go beyond this, and conclude that our embedding is a valid model for mental representations of objects -- insofar as our interactions with them go -- would require additional experiments. One possibility would be to explicitly ask human subjects "given objects that load highly on this embedding dimension, what can you do with them", and consider the typicality of verb answers against the weight given to those verbs by the dimension. Given that our embedding is based on language data about which verbs apply to which objects, we would expect these experiments to give verb loadings coherent with ours.

A future direction of work will be to predict SPoSE dimensions that are not well explained in terms of affordance embeddings. We plan to do this using embeddings produced with the same framework, but from different co-occurrence statistics. The first possibility will be to extract instances in corpora where objects are the {\em subjects} of verbs, i.e. they act or cause certain effects. The second possibility will be to consider applications of adjectives to objects, given that those may contain information relevant to all 4 types of SPoSE dimensions. Finally, we will consider reducing visual representations of objects obtained through deep neural networks to embedding vectors, as those contain both visual and semantic information.

%Another future direction will be to understand how affordance could be driven by more fine-grained visual appearance properties, by considering other semantic dependencies, or jointly using text and image features such as \cite{wang2020learning}.

%To increase prediction quality in future work, one approach will be to enrich and refine the co-occurrence matrix in larger corpora, now that the basic approach has been shown to be feasible. A

\subsubsection*{Acknowledgments}
This work was supported by the National Institute of Mental Health Intramural Research Program (ZIC-MH002968, ZIA-MH002035). This work utilized the computational resources of the NIH HPC Biowulf cluster (http://hpc.nih.gov). The authors would like to thank Martin Hebart and Charles Zheng for patiently sharing SPoSE and THINGS resources with us, and Aria Wang for graciously giving us access to her object affordance dataset.

\bibliographystyle{apacite}
\setlength{\bibleftmargin}{.125in}
\setlength{\bibindent}{-\bibleftmargin}
{\footnotesize \bibliography{cogsci2021}}

\appendix
\section{Appendix: Hyper-parameter Selection for Non-negative Matrix Factorization}
\label{appendix:hyperparameter}
\begin{figure}[h!]
    \begin{center}
    \includegraphics[width=1\linewidth]{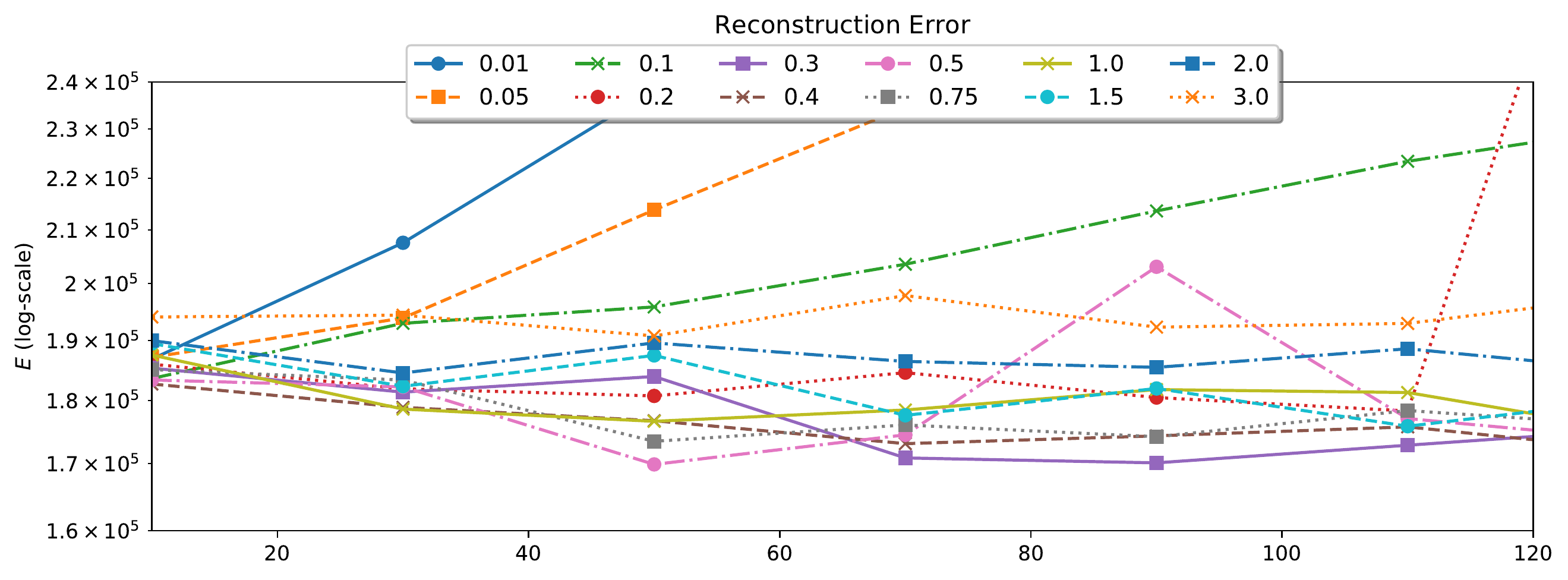}
    \end{center}
    \caption{A zoom-in plot for the reconstruction errors.}
    \label{fig:Hyperparameter}
\end{figure}
Denote $M_t, M_v \in \{0, 1\}^{n\times m}$ to be the mask matrices for indicating held-in and held-out entries of the input PPMI matrix $P$ in CV procedure, we then optimize for $O^*$ and $V^*$:
\begin{equation}
O^*, V^* = \underset{O, V}{\text{argmin}} \Vert M_t \odot (P - OV^T) \Vert_F^2 + \beta \mathcal{R}(O, V).
\end{equation}
To apply the multiplicative update scheme in~\cite{lee2001algorithms}, we need the partial derivatives with respect to $O$ and $V$. Denote $F(O, V) \equiv \Vert M_t \odot (P - OV^T) \Vert_F^2 + \beta \mathcal{R}(O, V)$, we have
\begin{equation}
\begin{split}
\nabla_O F(O, V) &= (M_t \odot OV^T) V - (M \odot P) V + \beta \cdot \mathbf{1}\\
\nabla_V F(O, V) &= (M_t \odot OV^T)^T U - (M \odot P)^T U + \beta \cdot \mathbf{1}.
\end{split}
\end{equation}
We then have the following update rules that is guaranteed to be non-increasing:
\begin{equation}
\begin{split}
    O^{(i+1)} &\leftarrow O^{(i)} \odot \frac{(M_t \odot P) V^{(i)}}{(M_t \odot O^{(i)}(V^{(i)})^T) V^{(i)} + \beta} \\
    V^{(i+1)} &\leftarrow V^{(i)} \odot \frac{(M_t \odot P)^T U^{(i)}}{(M_t \odot O^{(i)}(V^{(i)})^T)^T U^{(i)} + \beta},
\end{split}
\end{equation}
where the fraction here represents elementary-wise division. For the choice of $M_t$ and $M_v$, we follow the same approach as proposed in~\cite{kanagal2010rank}. We first split the matrix into $K$ blocks, with randomly shuffled rows and columns. Denote $\mathtt{r}^{(k)}$ and $\mathtt{c}^{(k)}$ to be the index vectors for rows and columns respectively, where $\mathtt{r}^{(k)}_i = 1$ if row $i$ is in block $k$, or $\mathtt{c}^{(k)}_j = 1$ if column $j$ is in block $k$. The mask for $k$-th block can then be expressed as $M^{(k)} = \mathtt{r}^{(k)} \otimes \mathtt{c}^{(k)}$. We then randomly select $q$ out of $K$ blocks as holdout blocks, which gives
\begin{equation}
    M_v = \sum_{s=1}^q \mathtt{r}^{(k_s)} \otimes \mathtt{c}^{(k_s)}, \quad M_t = \mathbf{1} - M_v,
\end{equation}
where $k_s$ is the index of selected block. The reconstruction error $E$ can thus be computed:
\begin{equation}
E = \Vert M_v \odot (P - O^*(V^*)^T) \Vert_F^2 + \beta \mathcal{R}(O^*, V^*).
\end{equation}
Figure~\ref{fig:Hyperparameter} shows a zoom-in plot of the reconstruction error under different combinations of $d$ and $\beta$. For every ($d$, $\beta$) setting, we perform multiple optimization since NMF is sensitive to initialization. We then choose $d = 70$ and $\beta = 0.3$ accordingly. Empirically, we observe that the rank selection is quite robust to over-fitting when there is a sufficient sparsity control, for instance, $\beta > 0.1$ in our dataset. We also observe that whenever $d \in [50, 150]$ and $\beta \in [0.05, 0.5]$, the results are similar.

\newpage 

\section{Appendix: Distribution of AAUCs on WTaction and MSCOCO Dataset}
\label{appendix:AAUC}
The following figures show the AAUC distribution of the top 5 embeddings on the WTaction and MSCOCO dataset respectively.

\begin{figure}[htb]
    \begin{center}
    \includegraphics[width=1\linewidth]{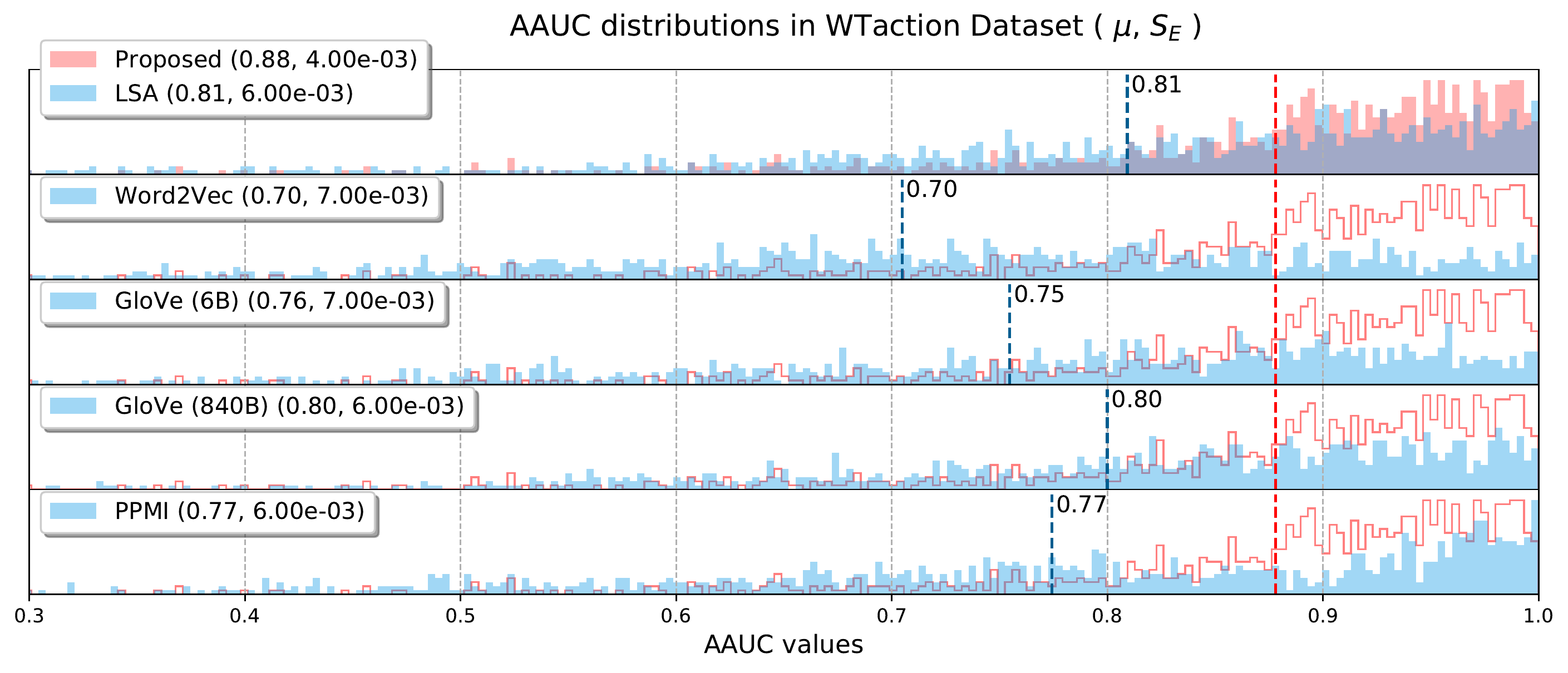}
    \end{center}
    \caption{AAUC Distribution on WTaction Dataset}
    \label{fig:Aria-AAUC}
\end{figure}

\begin{figure}[htb]
    \begin{center}
    \includegraphics[width=1\linewidth]{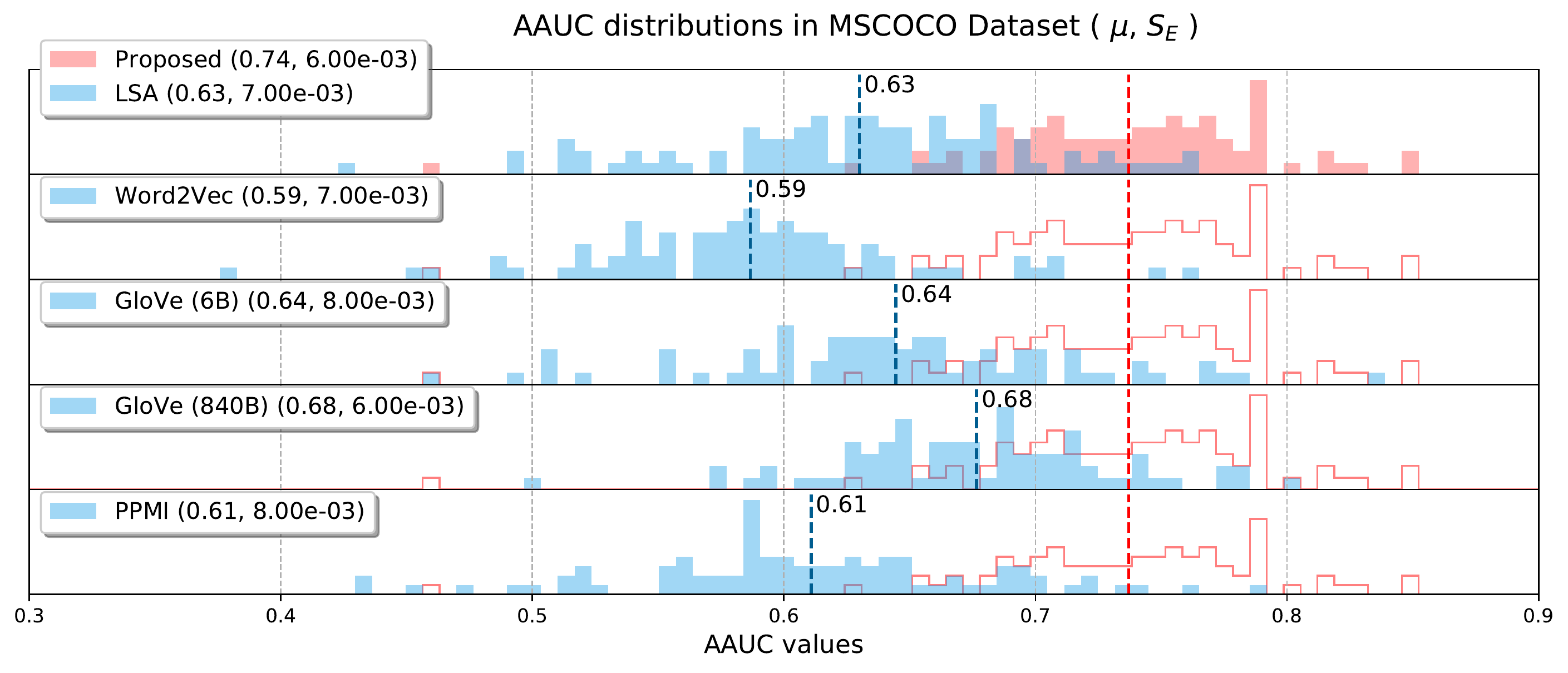}
    \end{center}
    \caption{AAUC Distribution on MSCOCO Dataset}
    \label{fig:Mihalcea-AAUC}
\end{figure}

\newpage
\section{Appendix: Prediction of SPoSE dimensions}
\label{appendix:SPoSE_pattern}
\begin{figure}[h!]
    \begin{center}
    \includegraphics[width=1\linewidth]{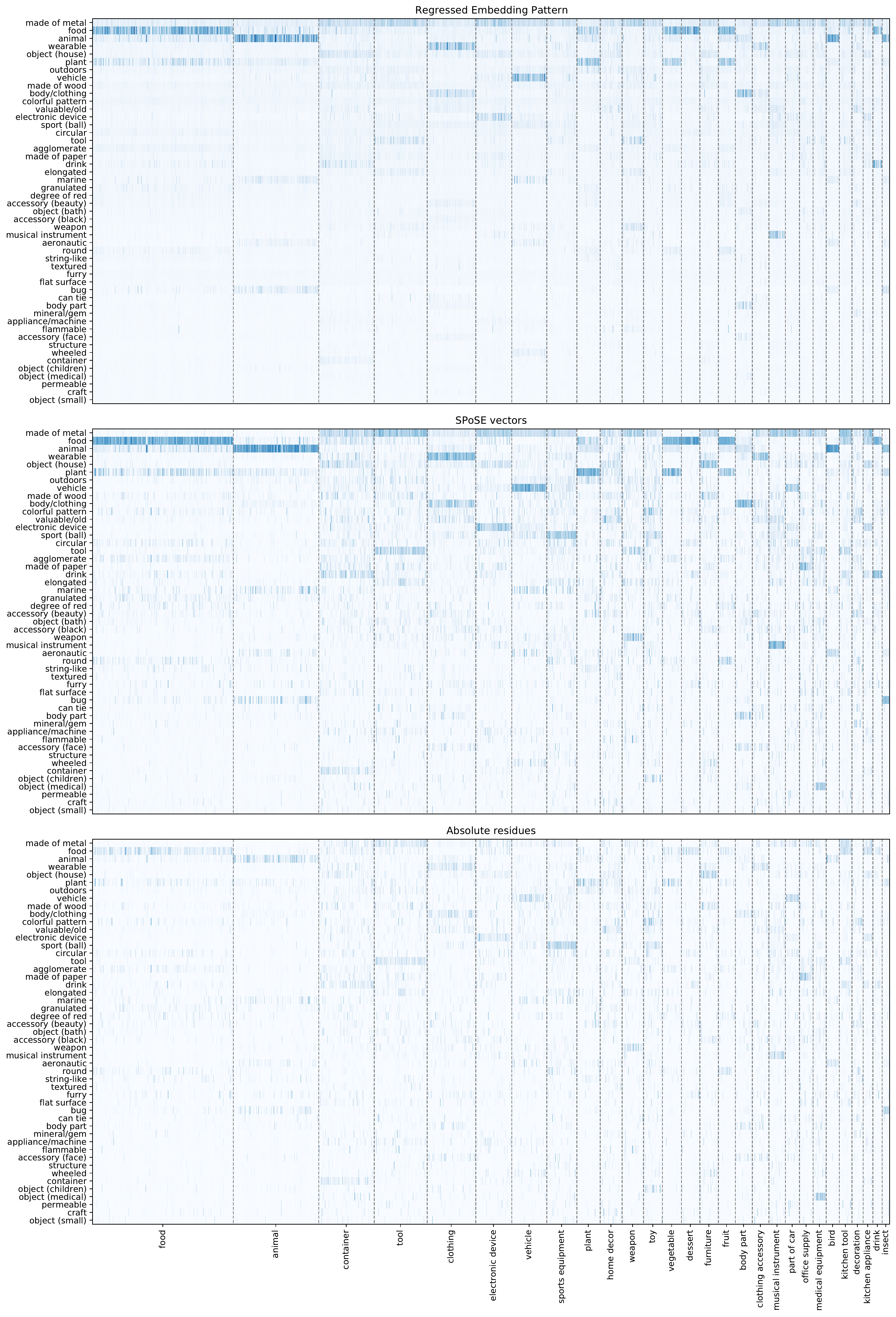}
    \end{center}
    \caption{Prediction of SPoSE embeddings from affordance embeddings (top) versus actual SPoSE embeddings (middle). Objects are grouped by semantic category (those with $\ge 15$ objects). The absolute residues of the prediction is also shown (bottom). Color range is fixed to show the magnitude of residues.}
    \label{fig:SPoSE_approximation_pattern_full}
\end{figure}

\newpage
\onecolumn
\section{Appendix: Affordances Assignment for each SPoSE Dimension}
\label{appendix:affordance_assignment}

\begin{table}[htb]
\tiny
    \begin{tabular}{rrlll}
    Pearson & & Dimension & & \\
    correlation & $p$-value & label & Taxonomy & 
    Affordances (Top Ten Ranked Verbs) \\
    \midrule
    
0.84 & 0.0e+00 & animal & categorical & kill, spot, hunt, observe, chase, feed, slaughter, sight, trap, find \\
0.82 & 0.0e+00 & food & categorical & serve, eat, cook, prepare, taste, consume, add, mix, stir, order \\
0.75 & 6.9e-323 & wearable & categorical & wear, don, match, knit, sew, fasten, rip, embroider, tear, model \\
0.75 & 9.7e-317 & vehicle & categorical & drive, hire, park, equip, rent, commandeer, crash, board, build, operate \\
0.71 & 2.0e-271 & plant & categorical & grow, cultivate, plant, add, eat, chop, gather, cut, dry, prune \\
0.68 & 8.6e-240 & musical instrument & categorical & hear, sound, play, learn, amplify, blare, study, tootle, toot, tinkle \\
0.68 & 2.6e-239 & electronic device & categorical & install, operate, connect, activate, invent, disconnect, manufacture, purchase, incorporate, design \\
0.67 & 2.5e-227 & made of metal & structural & fit, invent, manufacture, incorporate, design, position, attach, utilize, carry, install \\
0.66 & 6.2e-219 & body/clothing & categorical & wear, don, straighten, slash, bandage, hurt, rip, injure, heal, model \\
0.63 & 4.0e-196 & bug & categorical & kill, catch, spot, observe, find, eradicate, deter, trap, hunt, feed \\
0.63 & 8.9e-195 & drink & categorical & drink, pour, quaff, sip, guzzle, sup, swig, spill, imbibe, gulp \\
0.61 & 7.3e-183 & tool & categorical & wield, grab, hold, carry, sharpen, swing, hand, pick, clutch, throw \\
0.57 & 7.3e-152 & can tie & functional & fasten, tighten, unfasten, undo, attach, thread, tie, secure, loosen, loose \\
0.55 & 3.0e-142 & marine & categorical & spot, beach, moor, sail, observe, find, sight, capsize, catch, call \\
0.54 & 2.6e-134 & granulated & structural & contain, mix, scatter, add, gather, remove, sprinkle, dry, deposit, shovel \\
0.52 & 6.7e-125 & valuable/old & functional & steal, discover, find, carve, place, design, contain, recover, craft, hide \\
0.52 & 6.0e-124 & object (house) & categorical & design, manufacture, fit, incorporate, install, place, fill, purchase, clean, buy \\
0.52 & 1.3e-122 & aeronautic & categorical & spot, observe, sight, destroy, build, construct, find, photograph, equip, design \\
0.52 & 7.7e-121 & outdoors & categorical & remove, construct, place, surround, incorporate, carry, erect, design, fit, build \\
0.50 & 7.6e-114 & body part & categorical & sprain, fracture, bandage, flex, injure, rest, bruise, straighten, hurt, hyperextend \\
0.50 & 7.9e-114 & weapon & categorical & throw, carry, hurl, drop, grab, wield, retrieve, fire, hold, toss \\
0.48 & 3.2e-100 & flammable & functional & light, extinguish, ignite, throw, carry, flash, kindle, place, manufacture, douse \\
0.47 & 5.5e-96 & textured & appearance & remove, place, hang, tear, stain, spread, weave, clean, drape, wrap \\
0.44 & 1.3e-83 & round & appearance & grow, cultivate, pick, add, slice, place, eat, chop, throw, plant \\
0.44 & 4.7e-82 & accessory (beauty) & categorical & steal, wear, find, place, gather, pick, remove, contain, give, sell \\
0.42 & 1.1e-77 & mineral/gem & categorical & steal, discover, recover, contain, retrieve, find, hide, place, remove, incorporate \\
0.42 & 2.8e-74 & accessory (face) & categorical & wear, remove, don, place, design, buy, call, pull, bring, find \\
0.40 & 3.4e-68 & made of wood & structural & place, remove, carry, incorporate, design, contain, bring, construct, manufacture, find \\
0.40 & 4.5e-67 & container & functional & empty, fill, carry, place, clean, load, bring, dump, unload, leave \\
0.38 & 2.9e-61 & elongated & structural & grab, carry, wield, hold, pick, place, throw, hand, bring, drop \\
0.38 & 1.4e-60 & structure & structural & incorporate, construct, fit, design, erect, position, install, mount, place, build \\
0.38 & 3.8e-60 & object (bath) & categorical & manufacture, remove, place, invent, clean, buy, put, apply, contain, design \\
0.37 & 1.1e-56 & string-like & appearance & remove, cut, place, pull, wrap, attach, manufacture, contain, call, bring \\
0.36 & 3.1e-56 & wheeled & functional & drive, manufacture, hire, design, equip, fit, rent, park, purchase, invent \\
0.36 & 2.2e-55 & made of paper & structural & manufacture, design, purchase, buy, place, invent, introduce, incorporate, fit, sell \\
0.33 & 4.4e-47 & appliance/machine & categorical & fit, manufacture, connect, design, install, incorporate, attach, utilize, purchase, invent \\
0.32 & 1.6e-43 & sport (ball) & categorical & manufacture, design, invent, buy, purchase, grab, carry, introduce, fit, bring \\
0.32 & 4.3e-43 & object (children) & categorical & buy, manufacture, design, purchase, bring, find, sell, introduce, steal, call \\
0.31 & 2.5e-39 & agglomerate & appearance & place, contain, add, remove, sell, manufacture, combine, find, buy, steal \\
0.29 & 1.3e-34 & flat surface & appearance & place, bring, remove, put, grab, hang, wrap, manufacture, buy, pull \\
0.28 & 9.9e-33 & circular & appearance & place, incorporate, fit, invent, remove, manufacture, design, call, position, utilize \\
0.26 & 1.4e-28 & accessory (black) & categorical & remove, grab, manufacture, buy, place, design, bring, wear, carry, invent \\
0.24 & 2.0e-24 & colorful pattern & appearance & manufacture, buy, design, place, remove, sell, invent, purchase, contain, bring \\
0.23 & 4.0e-23 & object (medical) & categorical & invent, place, remove, find, contain, manufacture, design, bring, carry, buy \\
0.23 & 1.3e-21 & craft & appearance & place, bring, remove, design, hang, call, buy, put, pull, manufacture \\
0.22 & 1.9e-21 & permeable & structural & fit, incorporate, remove, place, design, manufacture, install, position, clean, attach \\
0.18 & 6.5e-14 & object (small) & categorical & place, call, remove, buy, find, bring, manufacture, introduce, contain, incorporate \\
0.18 & 8.9e-14 & degree of red & appearance & place, call, add, contain, remove, find, buy, bring, introduce, sell \\
0.12 & 7.6e-07 & furry & appearance & place, call, find, remove, buy, bring, introduce, contain, add, manufacture \\

    \end{tabular}
    \vspace{0.5em}
    \caption{Affordance assignment for SPoSE vectors ordered in terms of Pearson correlation with regression outcomes. The dimension labels are vastly simplified. The full descriptions for each dimension are provided in Appendix~\ref{appendix:SPoSE_description} (Table~\ref{table:full_descriptionA} and Table~\ref{table:full_descriptionB}). }
    \label{table:assignment_supplementary}
\end{table}

\clearpage

\onecolumn
\section{Appendix: Full descriptions for each SPoSE dimension}
\label{appendix:SPoSE_description}
The abbreviation and the full description of every SPoSE dimension.

\begin{table}[h!]
\scriptsize
\setlength{\extrarowheight}{3pt}
    \begin{tabular}{p{0.15\textwidth}L{0.8\textwidth}}
    Abbreviation & Full descriptions \\
    \midrule
accessory (black) & accessories, beauty, black, blackness, classy, date, emphasize, fancy, hair, hard, high-class, manly, objects, picture, telescope \\

accessory (face) & accessories, body parts, culture, decoration, eyes, face, face accessories, facial, goes on head, hair, head, less appealing, senses, touches face, wearable \\

accessory (beauty) & accessory, beautiful, beauty, color, fancyness, feminine, feminine items, floral, flowers, flowery, gentle, girly, love, muted  colors, pastel, pink \\

aeronautic & aero-nautic, air, airplanes, aviary, aviation, buoyant, flies, flight, fly, flying, flying to not, high in air, light, move, sky, swim, transportation, travel \\

agglomerate & accumulatable, bundles, collection, colors, countable, grainy to smooth, group of similar things, groupings, groups, groups of small objects, large groups, little bits of things, many, metals, nuts, objects, patterns, piles, quantity of objects in photo, round, small, small objects in groups, small parts that look alike, symmetrical \\

animal & animal, animals, animals in zoo, animals that do not fly, from complex to less, fuzzy, grass, ground animals, land animals, mammal, mammals, natural, size, wild animals, wild to human-made, wilderness \\

appliance/machine & building materials, construction, destructive, electric items, factory, farm tools, foundation, home tools, in groups, long, machinery, maintenance, mostly orange, processing, renovation, rocks, rope-like, thing, tool, tools \\

body part & body, body parts, esp extremities, extremities, extremities of body, feet, feet to hands, fingers, found on people, hand, hands, human, legs, limbs, lower body, skin \\

body/clothing & bodies, body, body accessories, body maintenance, body part- related, body parts, body parts with hair, face, how much skin showing, human, human body parts, part of body, parts, people, skin, touched by skin \\

bug & animals that stick onto things, ants, bug, bugs, can hurt you, dangerous, gardening, insects, interact with bugs, small animals, small to large, wild \\

can tie & bands, bondages, can tie, chained, circles, coils, construction, fasteners, knotting, long, rope, ropes, round, string-like, strings, tensile, thing can tie around, tied, ties, trapped, violent, wires, wrap, wrapped around to what gets wrapped \\

circular & circles, circular, cylindrical, discs, flat, round, shape, targets \\

colorful pattern & artistic, bright, bright colors, color, color variety, color vibrancy, colorful, colors, many colors, patterns \\

container & able to put something in it, boxes, buckets, can put things into, carts, container, containers, containers for liquids, containing, covering, cylinders, diverse, drums, enclosed objects, hold other things, hollow tubes, paints, shapes, storage, unknown \\

craft & a lot of patterns, art, artisinal, arts, candles, circles, color, crafts, detailed dots, do-it-yourself, grandma, grandparent-like, handmade, home  patterns, home-making related, housework, in grandma's home, intricacy, quilt, rectangles, sewing, specks, stitching, twine, unknown, weaving, wood, woven, yarn \\

degree of red & color, colors, degree of redness, red, red (bright) \\

drink & 3-dimensional, beverage, containers, containers for liquids, drinks, edible, glass, glasses, hold liquid, liquid, liquids, other things, things that fit in containers, things that hold liquids, vessels with liquids, viscosity \\

electronic device & digital, digital devices, digital media, electric, electronic, electronics, hard, hard to understand, media, old technology, technological, technology, telephones, typing instruments \\

elongated & able to be held, cane, cylinders, cylindrical, darts, grouped, long, long narrow, long objects, long-shaped, narrow, pen-like, pencils, pens, shape, sharp, skinny rectangles, stick-like, sticks, straight, straight to curved, symmetrical, thin \\

flammable & fire, flammability, flammable, heat, hot, light, outdoors, warm \\

flat surface & attaching, breakable, clean, cloth, convenience, disposable, flat coverings, gathering, grated pattern, handle everyday, helpful, hold things in, multi-shaped, not smooth to smooth, paper, paper-like, sheets, stick-like, thin, things that roll, tissue, white \\

food & baked food, baked goods, carbs, cheesy, comforting, cooked, deliciousness, edible, entrees, food, made dish, natural products, nutrients, pastry, prepared food, processed, salt, where it comes from \\

furry & fluffy, furry, more of one color, white, white and fluffy, winter \\

granulated & a lot of items, ash, color, dirt, elements, grain-looking, grains, grainy, grainyness, granular objects, ground, ground (grinded), homogeneity, lots of same, many, not colorful, particles, rocky, shape, size of particles, small, small particles, stones, tiny groupings, tone, unknown minerals or drugs \\

made of metal & buckle, build, building, gray, hard, metal, metal tools, metallic, metallic tools, metals, shiny, silver, tools, use with hands \\

made of paper & books, card, classroom, collections, flatness, found in office, groups, has text, note-taking, office, paper, paper (colorful), papers, printed on, reading materials, rectangles, school, square, squares, stacks, striped, work \\

made of wood & brown, made of wood, natural, natural resources, orange, wood, wood-colored, yellow \\

marine & aquamarine life, aquatic activities, cruise, fish, in water, marine, nautical, ocean, outdoor, outside water, paradise, sea, ships, vacation, water \\

    \end{tabular}
    \vspace{0.5em}
    \caption{Descriptions of abbreviation (A)\label{table:full_descriptionA}}
\end{table}

\begin{table}[t]
\scriptsize
    \setlength{\extrarowheight}{3pt}
    \begin{tabular}{p{0.15\textwidth}L{0.8\textwidth}}
    Abbreviation & Full descriptions \\
    \midrule
mineral/gem & beauty, clear minerals, crystal, earth-derived, gems, ice, in an artistic way, inspecting, intricate, jewelry, jewels, metallic, natural, natural minerals, prized, pure, rare, reflective, round, roundish, sharp, shiny, shinyness, sterile, translucent, valuable \\

musical instrument & control noise, hearing, instruments, listen, listening, loud, make noise, music, music instruments, musical, musical instruments, recreational instruments \\

object (bath) & bathroom, cleaning, essential everyday, gray, home  (inside home), household items, hygiene, self-care, soap, toiletries, water, white \\

object (children) & baby, baby toys, child, child-like, children, dolls, toys, young, youth \\

object (house) & bland to colorful, chairs, cloth, common in household, everyday household, flat, furniture, house, house essentials, house surfaces, household commonness, household furniture, in house, living furniture, main component of room \\

object (medical) & health, health-concerning, hospital, hygiene, injury, medical, medical instruments, medical supplies, medicine, sick, to good health, unhealthy, water, wellness \\

object (small) & ?, amish, appealing, candles, circular, color? Unsure, colorful, covers, cylinders, cylindrical, flat fat cylinders, hands-on, jewelry, lids, saving, shape, similar-shaped, things you grab, twine, unknown, yellow \\

outdoors & backyard, blue, brown colored, columns, common in outdoor, dirt, garden, landscape, man-made, monuments in nature, natural, nature, not colorful, outdoor objects, outdoorsy, park, pavement, pristine, public, quiet, rocks, rough, rows, scenery, separaters, stand on their own, statues, stone, tools, wood, woodsy, yard \\

permeable & can pass through, dot patterned, grates, holed, knit, little holes, mesh, metal, net, nets, octagonal, pattern, patterned, patterns, patterns (holes), repeated patterns, repeating, repeating patterns, repetitive, shiny, silver, small, strainer \\

plant & green, green leaves, green plants, green plants and herbs, greenery, greenness, greens, grow, natural, nature, plant-like, plants, things that grow from earth, vegetable, vegetables \\

round & artistic, ball, balls, circular, circular and colorful, circus, contrasting circles, fruits, kid, pictures, round, rund, shape, spherical \\

sport (ball) & athletic, ball  toys, balls) to less active, competition, recreation, round, sport, sports, sports (active, sporty \\

string-like & amount of rigid ends, confetti, different shapes, elongated, hay, high-density, knots, lines, lines jutting out, long, long things mashed up, look like sticks, mesh, netting, patterns, prickly, protruding, repeating in an ordered way, rope, ropes, skinny, spiky, stacks, strings, stringy, symmetrical, tangled, twirled around \\

structure & amusement, antenna, big, caged, common to humans, complete, disordered, electrical, elongated, enclosures, found outside, grand, high, high in air, industrial, ladder, large, multiple cylinders, multiple similar things, narrow, outdoor, part of circle, shapes, stacks, structural, tall, things that go up, things that hang, trapped, wires \\

textured & appealing, carpets, flat coverings, fractals, lay flat, mesh, pattern, patterned, patterns, pieces, rectangles, repeating shapes, repetitive, rugs, sheets, small repeating patterns, squares, textured \\

tool & elongated, hand tools, household, jagged, long, pointy tools, pole, saws, scrape, sharp, sharp tools, small instruments, straight, tools, use with hands, utility, wedges \\

valuable/old & English royalty, antiquity, bottles, bronze, fine things, gaudiness, gold, high-class, high-quality, history, important, jewelry, jewels, monarchy, old, ornaments, precious metals, pristine, royal, royalty, shiny, silver, trophy, valueable \\

vehicle & can be moved, car, cars, complex vehicles, construction, efficiency of transportation, fast, ground motorized vehicles, machinery, mobility, move, speed of movement, transportation, transportation vehicles, travel with, truck, vehicles, wheeled vehicles, wheels \\

weapon & black, danger, dangerous, equipment, masculine, military, negatively-associated, ornaments, risky, self-defense, somber, violence, violent, war, war-like, weaponry, weapons \\

wearable & accessories, blue, can wear or carry or put on, clothes, clothing, cotton clothing, covering body, shirt, things to wear, things you wear, touch body, touch person, utilities, warm clothes, wearable \\

wheeled & able to hold, bicycle, caged, can sit on, can stand or drive, carrier, chair, destinations, holds objects, light movement, mobility, motion, move someone, playful, round, thing, things with wheels, trapped, wagon, wheel, wheeled-structures, wheels \\

    \end{tabular}
    \vspace{0.5em}
    \caption{Descriptions of abbreviation (B)\label{table:full_descriptionB}}
\end{table}

\clearpage

\section{Appendix: Components of SPoSE dimension approximation}
\label{appendix:SPoSE_components}

\begin{figure}[h!]
    \begin{center}
    \includegraphics[width=0.75\linewidth]{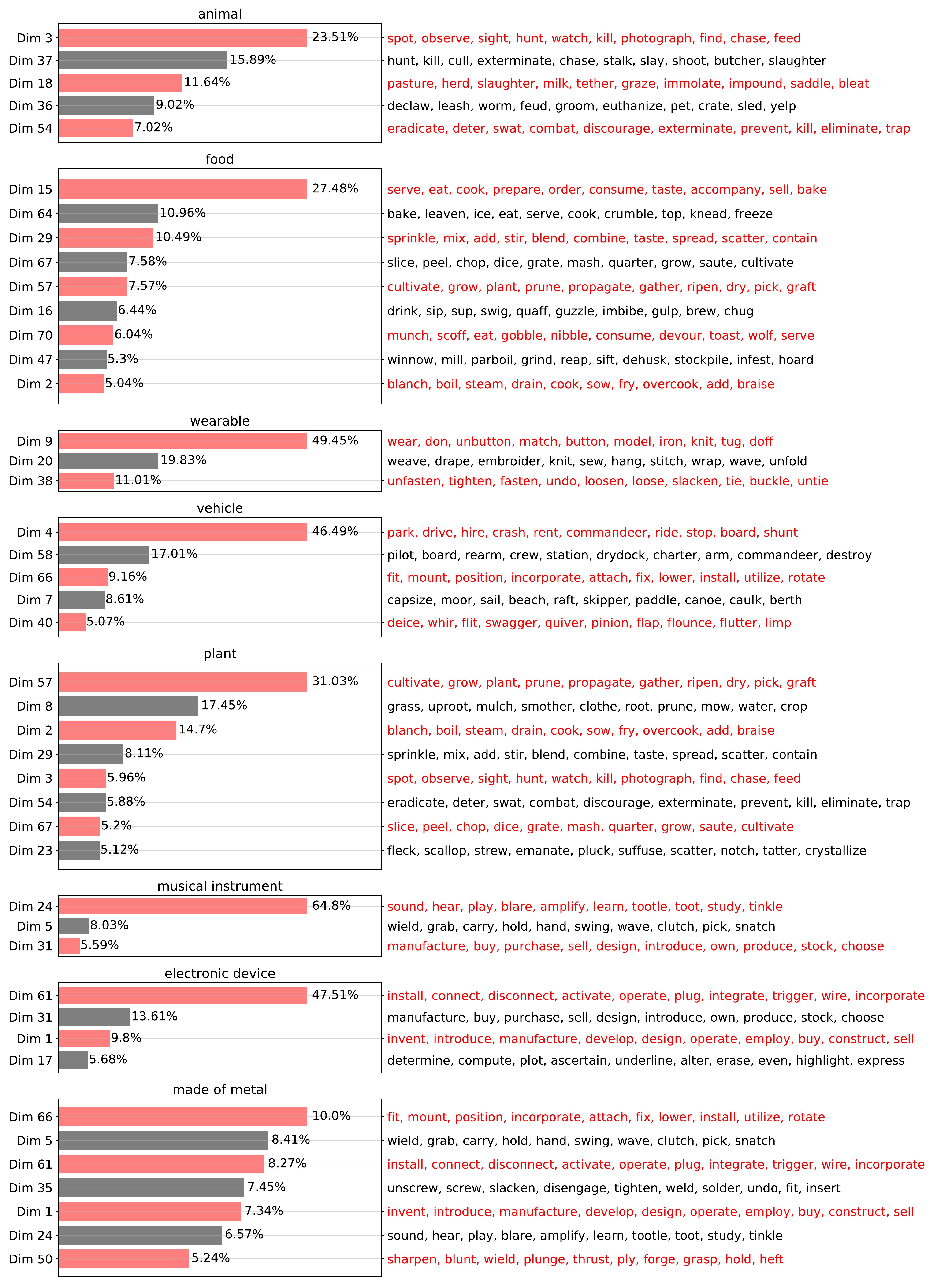}
    \end{center}
    \caption{Components of SPoSE dimension approximation (A)\label{fig:contributionA}}
\end{figure}

\begin{figure}[htb]
    \begin{center}
    \includegraphics[width=0.8\linewidth]{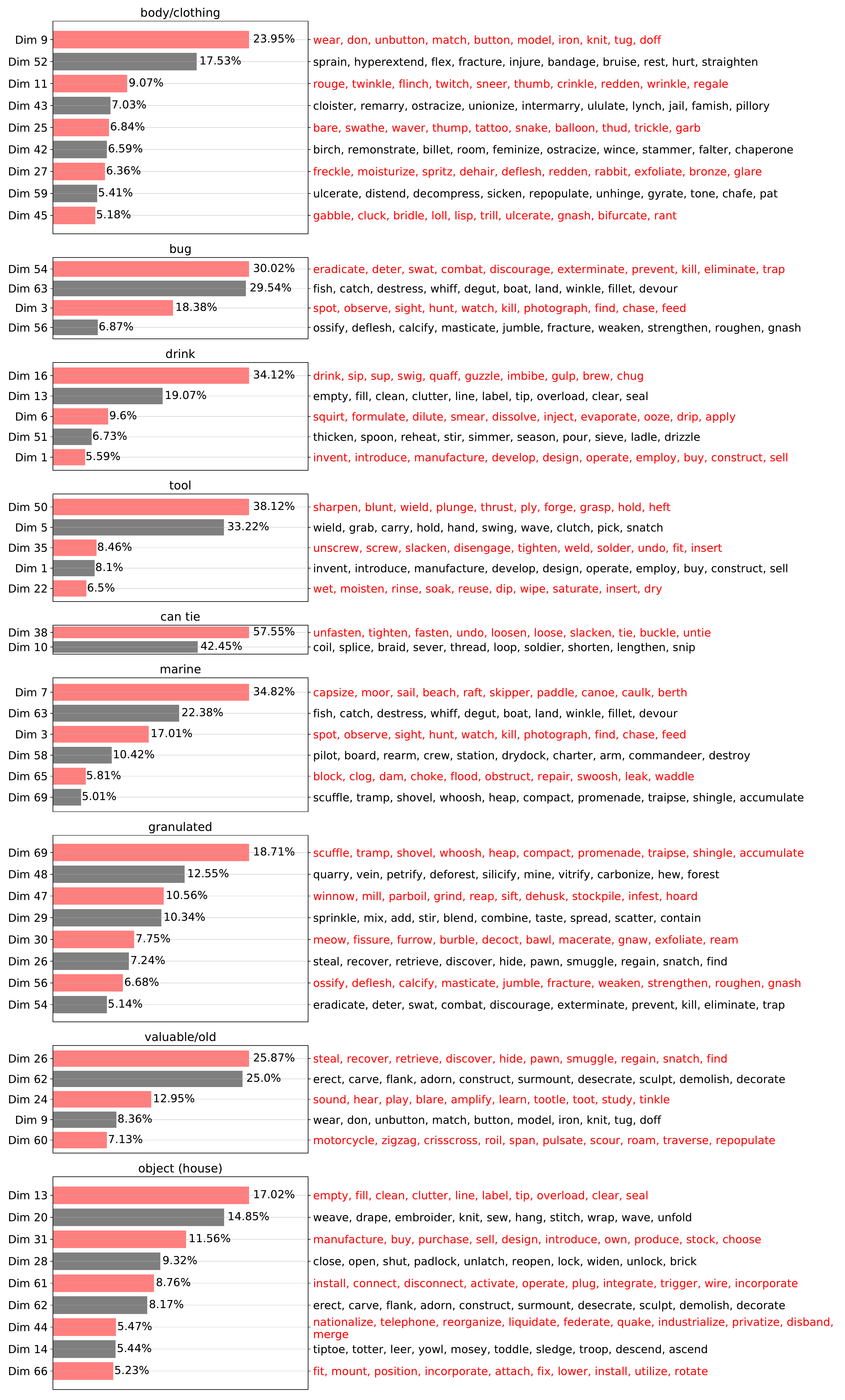}
    \end{center}
    \caption{Components of SPoSE dimension approximation (B)\label{fig:contributionB}}
\end{figure}

\begin{figure}[htb]
    \begin{center}
    \includegraphics[width=0.8\linewidth]{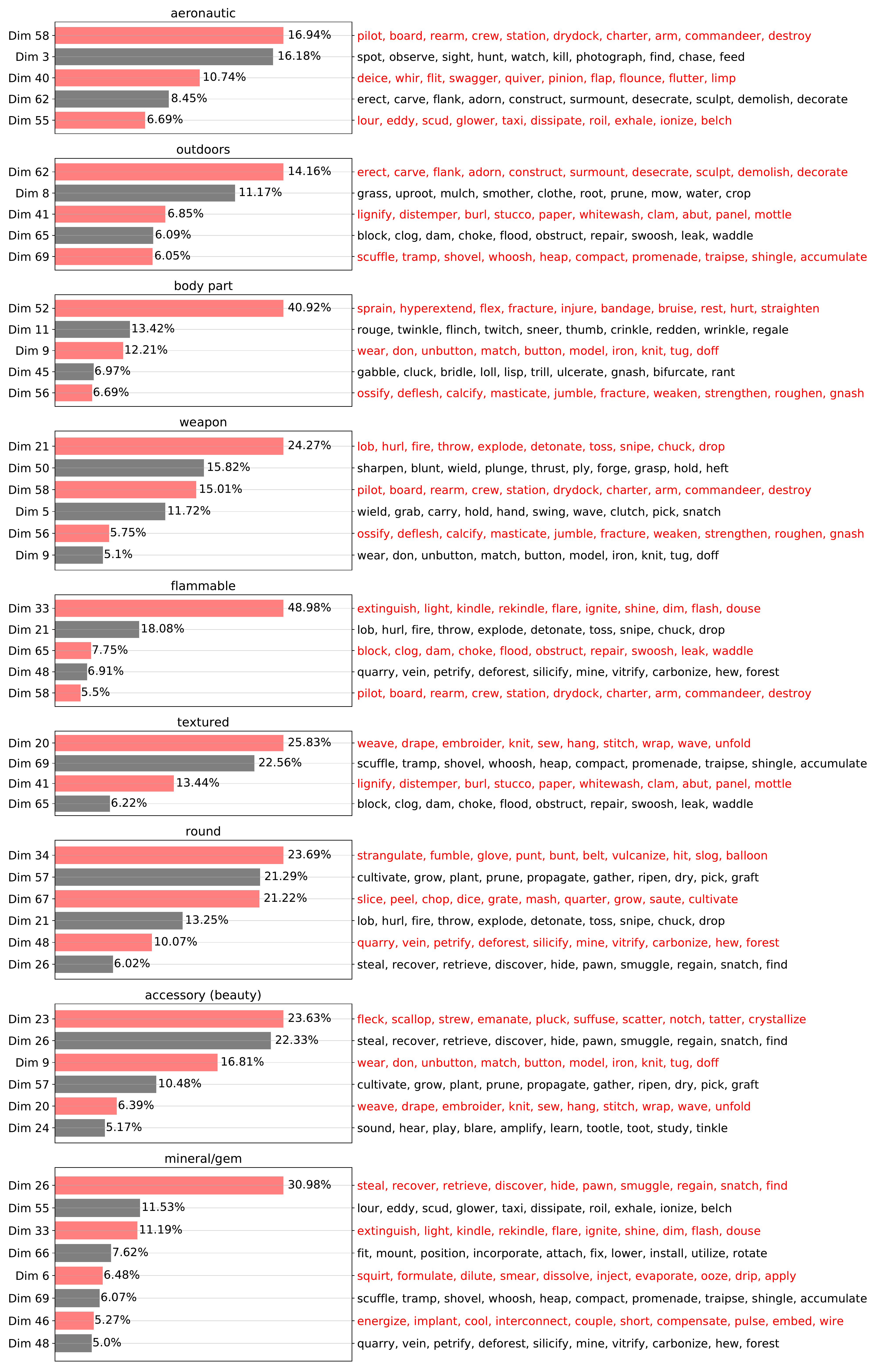}
    \end{center}
    \caption{Components of SPoSE dimension approximation (C)\label{fig:contributionC}}
\end{figure}

\begin{figure}[htb]
    \begin{center}
    \includegraphics[width=0.8\linewidth]{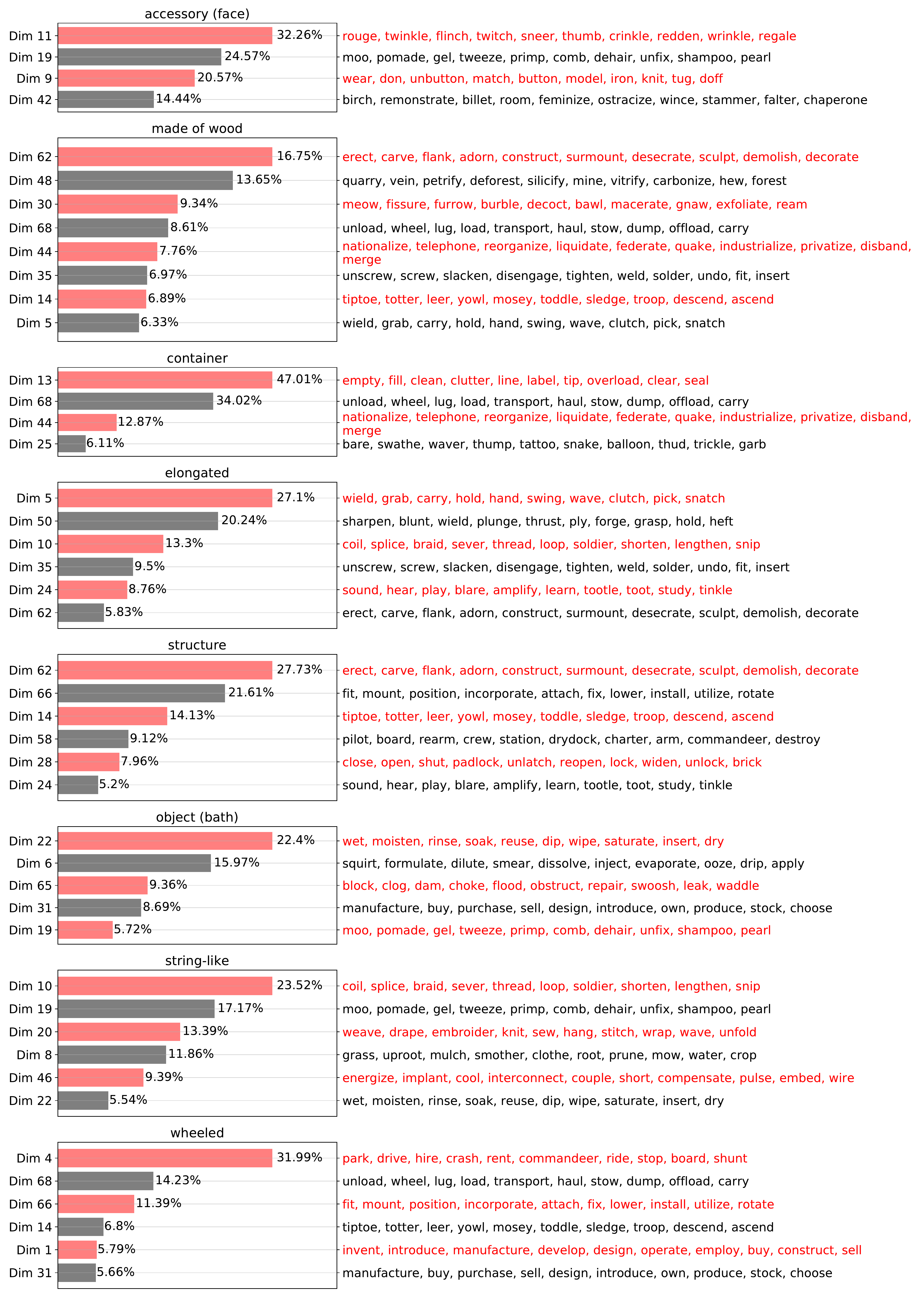}
    \end{center}
    \caption{Components of SPoSE dimension approximation (D)\label{fig:contributionD}}
\end{figure}

\begin{figure}[ht]
    \begin{center}
    \includegraphics[width=0.8\linewidth]{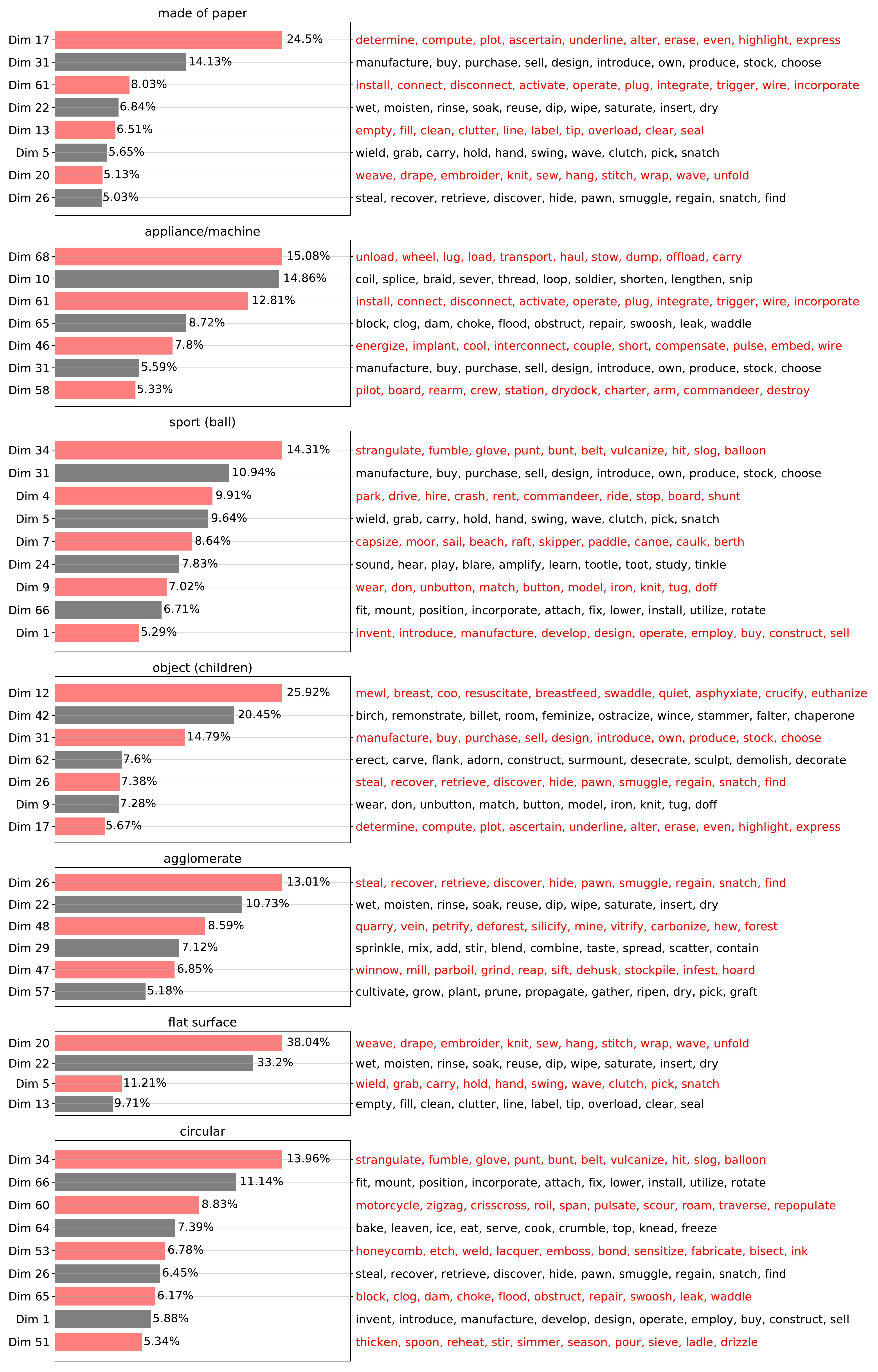}
    \end{center}
    \caption{Components of SPoSE dimension approximation (E)\label{fig:contributionE}}
\end{figure}

\begin{figure}[ht]
    \begin{center}
    \includegraphics[width=0.8\linewidth]{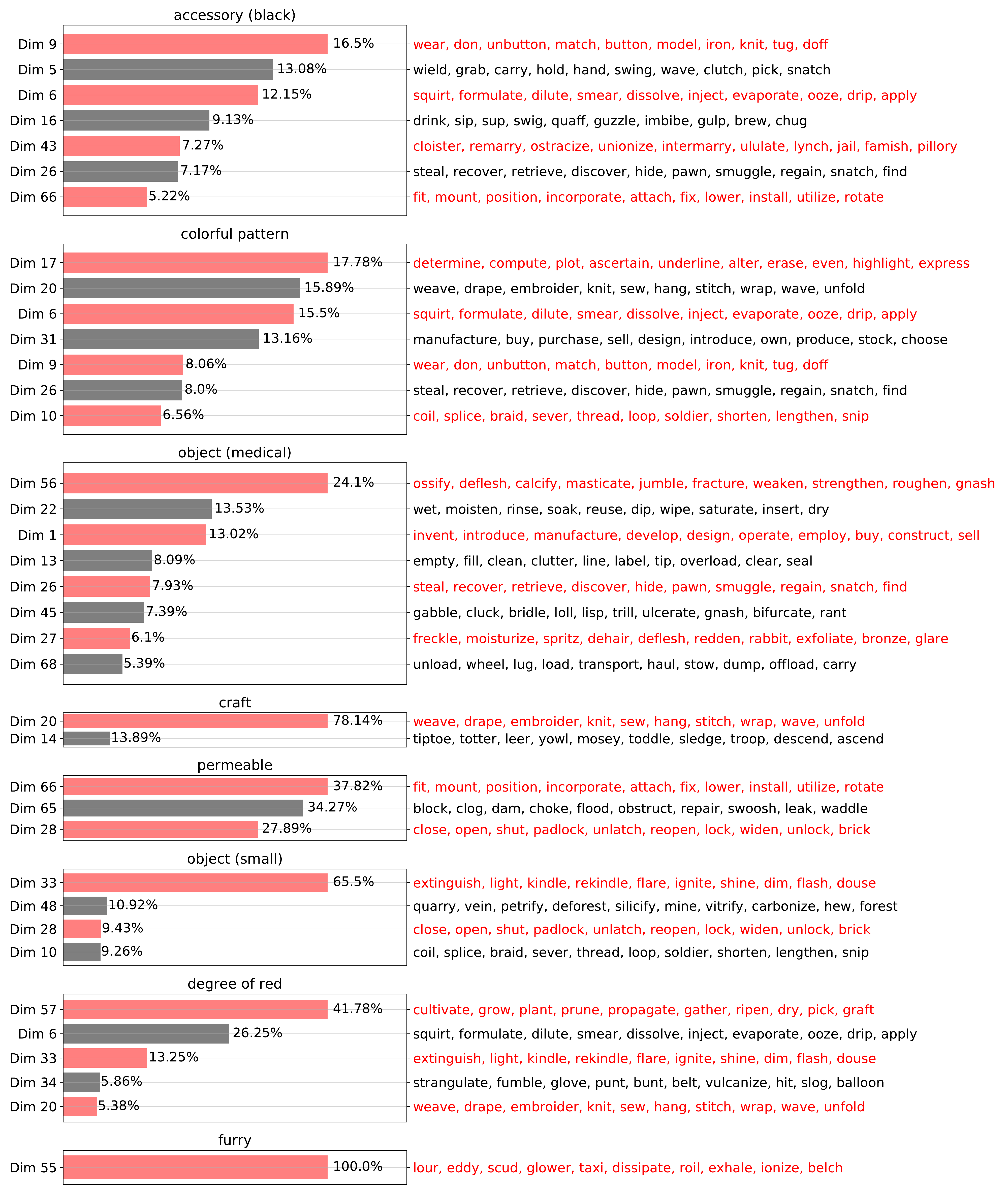}
    \end{center}
    \caption{Components of SPoSE dimension approximation (F)\label{fig:contributionF}}
\end{figure}

The following consecutive Figures~\ref{fig:contributionA}--\ref{fig:contributionF} show the component information of each SPoSE dimension. The percentage is calculated based on the portion $w_i \cdot \Vert O_{:,i}\Vert$, where $w_i$ is the regression coefficient corresponding to our $i$-th embedding dimension. The right ticks show the dimension affordances.

\end{document}